\documentclass[lettersize,journal]{IEEEtran}
\makeatletter
\DeclareRobustCommand\onedot{\futurelet\@let@token\@onedot}
\def\@onedot{\ifx\@let@token.\else.\null\fi\xspace}
\def\eg{\emph{e.g}\onedot} 
\def\ie{\emph{i.e}\onedot} 
 
 \def\vs{\emph{vs}\onedot}
\def\wrt{w.r.t\onedot~} 

\makeatother

\usepackage[dvipsnames,svgnames,x11names,table]{xcolor}
\usepackage[utf8]{inputenc} 
\usepackage[T1]{fontenc}    
\usepackage{url}            
\usepackage{booktabs,tabularx,colortbl,multirow,multicol,array,makecell,tabularray}
\usepackage{amsfonts}       
\usepackage{nicefrac}       
\usepackage{microtype}      
\usepackage{xcolor}         
\usepackage[linesnumbered,ruled,vlined]{algorithm2e}
\usepackage{algpseudocode}
\usepackage{times}
\usepackage{epsfig}
\usepackage{epigraph}
\usepackage{graphicx}
\usepackage{amsmath}
\usepackage{amssymb}
\usepackage{float}
\usepackage{soul}
\usepackage{pifont}
\usepackage{wrapfig}
\usepackage{mathtools}
\usepackage{xspace}
\usepackage{enumitem}
\usepackage{threeparttable}
\usepackage[pagebackref,breaklinks,colorlinks,citecolor=blue]{hyperref}

\usepackage{bm}
\usepackage{acronym}
\usepackage{xspace}
\usepackage{setspace}
\usepackage[capitalise]{cleveref}
\usepackage{overpic,wrapfig}
\usepackage[nocompress]{cite}
\usepackage[misc]{ifsym}
\usepackage{cite}
\usepackage{fontawesome}
\usepackage{float}
\usepackage{arydshln}
\usepackage{anyfontsize}

\acrodef{rl}[RL]{Reinforcement Learning}
\acrodef{il}[IL]{Imitation Learning}
\acrodef{slam}[SLAM]{simultaneous localization and mapping}
\acrodef{tamp}[TAMP]{task and motion planning}
\acrodef{urdf}[URDF]{unified robot description format}
\acrodef{sdf}[SDF]{Signed Distance Field}
\acrodef{pt}[\emph{pt}]{parse tree}
\acrodef{iou}[IoU]{intersection over union}
\acrodef{map}[mAP]{mean average precision}
\acrodef{vkc}[VKC]{Virtual Kinematic Chain}
\acrodef{sparc}[SPARC]{Spectral Arc Length}
\acrodef{or}[OR]{overlap ratio}
\acrodef{cd}[c. dist.]{closest distance}
\acrodef{bc}[BC]{Behavior Cloning}
\acrodef{vkc}[VKC]{Virtual Kinematic Chain}
\acrodef{eai}[EAI]{Embodied Artificial Intelligence}

\newcommand{\model}{\text{M$^2$Diffuser}\xspace}
\newcommand{\mpinets}{\text{M$\pi$Nets}\xspace}
\newcommand{\mpiformer}{\text{M$\pi$Former}\xspace}

\newcommand{\redtext}[1]{\textcolor{red}{#1}}

\let\oldnl\nl
\newcommand{\nonl}{\renewcommand{\nl}{\let\nl\oldnl}}
\newcolumntype{x}{>{\columncolor{MistyRose}}c}
\newcolumntype{y}{>{\columncolor{LightCyan1}}c}
\newcolumntype{z}{>{\columncolor{LightOliveGreen}}c}

\definecolor{grasping}{RGB}{179,0,0}
\definecolor{placement}{RGB}{70,130,180}
\definecolor{goal_reaching}{RGB}{143,179,0}
\definecolor{gpurple}{RGB}{128, 0, 128}
\colorlet{LightOliveGreen}{OliveGreen!15}

\makeatletter
\renewcommand{\paragraph}{%
  \@startsection{paragraph}{4}%
  {\z@}{0ex \@plus 0ex \@minus 0ex}{0em}%
  {\hskip\parindent\normalfont\normalsize\bfseries}%
}
\makeatother

\crefname{algorithm}{Alg.}{Algs.}
\Crefname{algocf}{Algorithm}{Algorithms}
\crefname{section}{Sec.}{Secs.}
\Crefname{section}{Section}{Sections}
\crefname{table}{Tab.}{Tabs.}
\Crefname{table}{Table}{Tables}
\crefname{figure}{Fig.}{Figs.}
\Crefname{figure}{Figure}{Figures}
\crefname{equation}{Eq.}{Eqs.}
\Crefname{equation}{Equation}{Equations}
\crefname{appendix}{Appx.}{Appxs.}
\Crefname{appendix}{Appendix}{Appendices}

\crefformat{equation}{Eq.~#2\textup{#1}#3}

\definecolor{gblue}{HTML}{4285F4}
\definecolor{gred}{HTML}{DB4437}
\definecolor{ggreen}{HTML}{0F9D58}

\ifCLASSOPTIONcompsoc
  \usepackage[nocompress]{cite}
\else
  \usepackage{cite}
\fi
\ifCLASSINFOpdf
\else
\fi
\begin{document}

%
\title {\model: Diffusion-based Trajectory Optimization for Mobile Manipulation in 3D Scenes
}
%
%
%
%

\author{
    Sixu Yan, 
    Zeyu Zhang,~\IEEEmembership{Member,~IEEE}, 
    Muzhi Han, 
    Zaijin Wang,
    Qi Xie, 
    Zhitian Li, 
    Zhehan Li, \vspace{3pt}\\
    Hangxin Liu,~\IEEEmembership{Member,~IEEE},
    Xinggang Wang,~\IEEEmembership{Senior Member,~IEEE}, and 
    Song-Chun Zhu,~\IEEEmembership{Fellow,~IEEE}

\IEEEcompsocitemizethanks{
\IEEEcompsocthanksitem {The work was done when Sixu Yan interned at BIGAI. \it{(Corresponding authors: Xinggang Wang and Hangxin Liu.)}}
\IEEEcompsocthanksitem{Sixu Yan and Xinggang Wang are with the School of Electronic Information and Communications, Huazhong University of Science and Technology, Wuhan 430074, China (e-mail: \href{mailto:yansixu@hust.edu.cn}{\textcolor{black}{{yansixu@hust.edu.cn}}}; \href{mailto:xgwang@hust.edu.cn}{\textcolor{black}{{xgwang@hust.edu.cn}}}).}
\IEEEcompsocthanksitem {Zeyu Zhang, Zaijin Wang, Qi Xie, Hangxin Liu and Song-Chun Zhu are with the State Key Laboratory of General Artificial Intelligence, Beijing Institute for General Artificial Intelligence (BIGAI), Beijing 100080, China (e-mail: \href{mailto:zhangzeyu@bigai.ai}{\textcolor{black}{{zhangzeyu@bigai.ai}}}; \href{mailto:wangzaijin@bigai.ai}{\textcolor{black}{{wangzaijin@bigai.ai}}}; \href{mailto:xieqi@bigai.ai}{\textcolor{black}{{xieqi@bigai.ai}}}; \href{mailto:liuhx@bigai.ai}{\textcolor{black}{{liuhx@bigai.ai}}}; \href{mailto:sczhu@bigai.ai}{\textcolor{black}{{sczhu@bigai.ai}}}).}
\IEEEcompsocthanksitem {Muzhi Han is with the Center for Vision, Cognition, Learning, and Autonomy (VCLA), Statistics Department, University of California, Los Angeles (UCLA), USA (e-mail: \href{mailto:muzhihan@ucla.edu}{\textcolor{black}{{muzhihan@ucla.edu}}}).}
\IEEEcompsocthanksitem {Zhitian Li is with the School of Automation Science and Electrical Engineering, Beihang University, Beijing 100191, China, and also interned at BIGAI (e-mail: \href{mailto:lizhitian1998@buaa.edu.cn}{\textcolor{black}{{lizhitian1998@buaa.edu.cn}}}).}
\IEEEcompsocthanksitem {Zhehan Li is with the School of Artificial Intelligence, Xidian University, Xi'an 710126, China, and also interned at BIGAI (e-mail: \href{mailto:zhehanli_robot@stu.xidian.edu.cn}{\textcolor{black}{{zhehanli\_robot@stu.xidian.edu.cn}}}).}
\IEEEcompsocthanksitem {Song-Chun Zhu is also with the School of Artificial Intelligence and the Institute for Artificial Intelligence, Peking University, Beijing 100871, China.}
}
}

\makeatletter
\g@addto@macro\@maketitle{
    \begin{figure}[H]
        \setlength{\linewidth}{\textwidth}
        \setlength{\hsize}{\textwidth}
        \centering
        \setcounter{figure}{0} 
        \includegraphics[width=\linewidth]{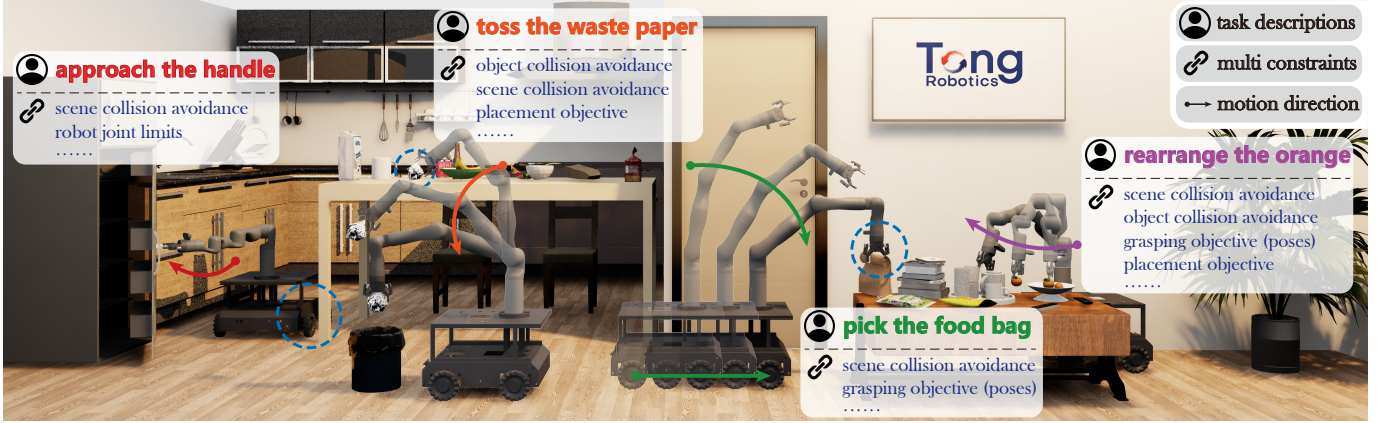}
        \vspace{-16pt}
        \caption{Challenges in mobile manipulation. Mobile manipulation plays a pivotal role in core robotic tasks such as object grasping, placement, rearrangement, and articulated object manipulation. Each of these tasks poses distinct challenges, requiring the motion generator to holistically account for constraints posed by environmental geometry and robot embodiment to accomplish task-specific objectives.}
        \label{fig:teaser}
    \end{figure}
    \vspace{-26pt}
}
\makeatother  
\maketitle

\IEEEtitleabstractindextext{%
\begin{abstract}
Recent advances in diffusion models have opened new avenues for research into embodied AI agents and robotics. Despite significant achievements in complex robotic locomotion and skills, mobile manipulation—a capability that requires the coordination of navigation and manipulation—remains a challenge for generative AI techniques. This is primarily due to the high-dimensional action space, extended motion trajectories, and interactions with the surrounding environment. In this paper, we introduce \model, a diffusion-based, scene-conditioned generative model that directly generates coordinated and efficient whole-body motion trajectories for mobile manipulation based on robot-centric 3D scans. \model first learns trajectory-level distributions from mobile manipulation trajectories provided by an expert planner. Crucially, it incorporates an optimization module that can flexibly accommodate physical constraints and task objectives, modeled as cost and energy functions, during the inference process. This enables the reduction of physical violations and execution errors at each denoising step in a fully differentiable manner. Through benchmarking on three types of mobile manipulation tasks across over 20 scenes, we demonstrate that \model outperforms state-of-the-art neural planners and successfully transfers the generated trajectories to a real-world robot. Our evaluations underscore the potential of generative AI to enhance the generalization of traditional planning and learning-based robotic methods, while also highlighting the critical role of enforcing physical constraints for safe and robust execution. Videos, code and more details are available at \url{https://m2diffuser.github.io}.

\end{abstract}

\begin{IEEEkeywords}
Mobile Manipulation, Embodied AI, Diffusion Model, Trajectory Generation and Optimization
\end{IEEEkeywords}}

\IEEEdisplaynontitleabstractindextext

%
\IEEEpeerreviewmaketitle

\section{Introduction}
\label{sec:introduction}
\IEEEPARstart{R}{esearch} into \ac{eai} increasingly emphasizes interaction with the environment, progressing from passive observation in learning visual navigation~\cite{gupta2017cognitive,zhu2017target} to active manipulation in object rearrangement~\cite{batra2020rearrangement,gu2022multi}, and more recently, to integrate large foundation models to tackle highly interactive tasks~\cite{firoozi2023foundation,yang2023foundation,huang2024rekep,han2024interpret,shang2024theia}. However, mobile manipulation~\cite{khatib1999mobile}—a core capability enabling agents to perform a wide range of tasks across large spaces—remains challenging for \ac{eai} agents.

The key difficulty in solving mobile manipulation tasks is the need to jointly account for agent embodiment, large-scale environment geometry, and task-specific objectives and constraints. For instance, as illustrated in \cref{fig:teaser}, when approaching and grasping an object, the success of the agent's motion execution depends not only on its own configuration but also on the state of its surroundings along its movements. Furthermore, even when picking the same object, the variations in context require task objectives tailored to ensure that the agent’s base position allows its arm to reach the object, the arm can avoid collisions with the environment, and the end effector can achieve a specific pose to execute the desired grasp, which all eventually lead to various types of motion constraints for the agent. However, the interdependencies—and sometimes conflicts—among these constraints present significant challenges for both traditional planning-based methods and learning-based approaches to solve the corresponding motion generation problems.

Encoding task objectives and related motion constraints implicitly in demonstration data or carefully designed reward or loss functions~\cite{tai2017virtual,li2020hrl4in,feng2021collision,fishman2022mpinets}, \ac{il} and \ac{rl} are usually used for \ac{eai} agents to learn sophisticated skills~\cite{lee2019composing,lee2020learning,lee2021adversarial,wu2023m, ni2023towards,wu2023tidybot,xiong2024adaptive}, complex locomotion~\cite{xia2021relmogen,gu2022multi, ma2022combining,sun2022fully,yokoyama2023asc}, whole-body motion~\cite{hu2023causal,fu2023deep,liu2024visual}, or advanced policies for long-horizon tasks~\cite{honerkamp2021learning,Honerkamp2023n2m2,Naik2024pregrasp}. However, they often struggle to fully eliminate violations of physical constraints in model inference. Moreover, expensive new data collection and model re-training are typically required to incorporate new task requirements. On the other hand, the field of robotics has a long history of developing planning and control methods to ensure robust and efficient executions for robots, a more concrete form of \ac{eai} agents. With substantial modeling techniques, various constraints can be formulated for robots to accomplish complex mobile manipulation tasks through whole-body control~\cite{minniti2019whole,stuede2019door,chiu2022collisionfreempc} and base-arm coordination~\cite{jiao2021consolidating,jiao2021efficient,li2024dynamic}. However, these approaches heavily rely on perfect knowledge of the environment~\cite{han2022scene,zhang2023part} and engineered goal proposals (\eg, grasp poses)~\cite{garrett2021integrated,marcucci2023motion}, limiting their scalability in real-world deployments. 

Recently, generative AI has demonstrated the remarkable ability to produce diverse and even novel content in text~\cite{liu2024intelligent}, images~\cite{rombach2022high,zhang2023adding}, and videos~\cite{blattmann2023align,blattmann2023stable} with high spatial and temporal consistency. However, the success of generative AI techniques has not yet been demonstrated in complex robotic tasks like mobile manipulation, primarily due to (i) the high dimensionality of the solution space, which requires efficient modeling and high-quality training data, and (ii) the strict requirement for physically precise execution, which demands high-fidelity model outputs. 

In this paper, we explore leveraging generative modeling techniques to produce holistic mobile manipulation motion that not only coordinates navigation and manipulation for obstacle avoidance, but also strictly satisfies task objectives with high precision, such as grasping objects. Specifically, we propose the \underline{M}obile \underline{M}anipulation \underline{Diffuser} (\model), a scene-conditioned diffusion model that takes robot-centric 3D scans to generate whole-body coordinated mobile manipulation motion. With the guided sampling mechanism inherent to diffusion models, \model incorporates explicit physical constraints (\eg, joint limits, scene collision and motion smoothness), as well as implicit task objectives (\eg, grasping pose selection), as cost and energy functions during its inference process. These functions are differentiable, effectively guiding the optimization of sampled trajectories to reduce physical violations and execution errors.

To develop \model, we first collect high-quality training data in simulated scenes by using an expert motion planner to generate whole-body mobile manipulation trajectories with smooth base-arm coordination. Second, we train \model by using robot-centric 3D scans, \ie, local point clouds represented in the robot's base coordinate, as model conditioning to improve scalability. By evaluating \model in both a physics-based simulator and real-world environments, we demonstrate that \model significantly outperforms state-of-the-art neural motion planners in generating robot motion for mobile manipulation tasks, where the robot must navigate toward and grasp 15 types of target objects. Furthermore, we show that the architecture of \model is flexible enough to be adopted to new mobile manipulation tasks, such as object placement and goal reaching. Our findings reveal two key insights for generative AI and \ac{eai}: (i) For motion generation, diffusion-based models offer a promising alternative to classical motion planning approaches, which require extensive prior knowledge and manual design, as well as to learning-based autoregressive planning approaches, which struggle to ensure physical safety and precise executions; and (ii) Even SOTA generative AI techniques, when trained with high-quality data from expert mobile manipulation planners, are still insufficient to guarantee safe execution. Effective enforcement of physical constraints during the generation process is critical for success in complex robotic applications.

\subsection{Related Work}
\paragraph*{Motion Generation in 3D Scenes} 
Generating robot motion requires understanding object geometry and scene context. Various 3D representations can be derived from raw observations, such as point clouds~\cite{fishman2022mpinets, Ze2024DP3, 3d_diffuser_actor, dalal2024neural}, voxels~\cite{johnson2020dynamically, shridhar2022peract, Ze2023GNFactor, huang2023voxposer, Ze2024DP3}, and implicit fields~\cite{yan2024dnact}, which support the learning of dexterous skills~\cite{Ze2024DP3} and object manipulation~\cite{shridhar2022peract, Ze2023GNFactor, goyal2023rvt, 3d_diffuser_actor, yan2024dnact}. Similarly, neural planners have been developed to imitate expert planner behaviors based on 3D representations of the environment~\cite{johnson2020dynamically, fishman2022mpinets, dalal2024neural}. While these methods accelerate motion generation, they primarily model robot motion generation as an autoregressive process, which struggles to capture complex trajectory distribution. In this work, the proposed \model learns trajectory-level distributions directly through a diffusion process, mitigating the weakness of existing methods in learning high-dimensional trajectory generation for mobile manipulation.
\paragraph*{Diffusion Models in Robotics} 
Due to their advantages in modeling multi-modal data distributions with stable training, diffusion models have been widely applied to robotic tasks like stationary manipulation~\cite{ajay2022conditional, janner2022planning, liu2023structdiffusion, chi2023diffusionpolicy, kapelyukh2023dall, carvalho2023motion, huang2023diffusion, urain2023se, mishra2023reorientdiff, yang2023policy, Ze2024DP3, 3d_diffuser_actor, ma2024hierarchical, yan2024dnact, kim2024robust}, autonomous navigation~\cite{sridhar2024nomad, song2024tgs, yu2024ldp}, quadruped locomotion~\cite{stamatopoulou2024dippest}, drone flight~\cite{das2024dronediffusion}, dexterous manipulation~\cite{zhang2024dexgrasp,weng2024dexdiffuser,yamada2024d}, and mechanical structures~\cite{wang2023diffusebot,xu2024dynamics}. However, applying diffusion models to mobile manipulation tasks requires high-quality training data, which is difficult to obtain. In this work, we leverage an expert planner from our previous research~\cite{jiao2021consolidating,jiao2021efficient} to collect a large set of whole-body mobile manipulation trajectories and use them to unlock the capability of diffusion models in complex robotic tasks.
\paragraph*{Learning-based Trajectory Optimization}
Trajectory optimization~\cite{betts1998survey, kelly2017introduction} has enabled robots to generate smooth and efficient movements. However, the requirement to explicitly define the objective and constraints limits the scalability of these techniques in complex tasks and challenging real-world environments. Existing work has addressed this limitation by applying deep learning to: (i) learn task goals such as grasp poses~\cite{sundermeyer2021contact, mousavian2019graspnet} and object affordance~\cite{mo2022o2o, noh2024learning,huang2023voxposer}, (ii) learn implicit objectives and constraint functions for grasping~\cite{urain2023se, mousavian2019graspnet, lu2020planning} and collision avoidance~\cite{danielczuk2021object, urain2022learning, koptev2022neural}, and (iii) learn neural motion planners from demonstrations for generating collision-free~\cite{fishman2022mpinets, huang2023diffusion} or kinematically feasible motion~\cite{fishman2022mpinets, huang2023diffusion, chi2023diffusionpolicy, zhu2023diffusion}. In this work, \model draws inspiration from these efforts and aims to tackle mobile manipulation. It directly learns trajectory-level distributions from expert demonstrations, followed by guided trajectory optimization with differentiable cost functions of physical constraints and task-oriented energy functions, which can be either explicitly defined or learned from data.
\paragraph*{Scene Understanding in Mobile Manipulation} 
Existing \ac{il}- and \ac{rl}-based household mobile manipulation primarily relies on egocentric or overhead observations, including RGB~\cite{wu2023tidybot}, RGB-D~\cite{wong2022error,xiong2024adaptive, mittal2022articulated}, and depth~\cite{ni2023towards, huang2023skill,gu2022multi} images. This setup provides only partial visual input, insufficient for capturing object-geometric and scene-spatial information, leading to poor understanding of constraints (\eg, occluded obstacles) between the robot embodiment and the environment. In this work, \model uses robot-centric 3D scans as visual input. Such 3D scans encode detailed geometric and spatial relationships around the robot, enhancing the model's understanding for physical constraints and task objectives. This design benefits decision-making and safe motion execution in cluttered 3D environments.

\subsection{Contribution}
To our knowledge, ours is the first work that applies diffusion models to solve robotic mobile manipulation tasks. It makes four major contributions:

\begin{enumerate}
\item We propose \model, the first scene-conditioned motion generator tailored for mobile manipulation in \ac{eai}. It seamlessly integrates multiple physical constraints and flexibly handles different task objectives, and directly generates highly coordinated whole-body motion trajectories with physical plausibility from 3D scans.

\item We highlight the importance of integrating physical constraints and task objectives into the generative process via a guided optimization mechanism, which ensures physical plausibility and task completion of the generated motion.

\item We demonstrate that the diffusion-based planner, compared to previous autoregressive planners, is better suited for generating high-dimensional mobile manipulation motion. It ensures spatial and temporal consistency of the generated trajectories.

\item We also show that taking local 3D scans around the robot as model visual input can be more effective for real-world generalization and deployment. 
\end{enumerate}

\subsection{Overview}
The remainder of this paper is organized as follows. In \cref{sec:m2diffuser}, we define the problem and provide a detailed introduction to our method. \cref{sec:sim_exp} presents the experimental setup and compares \model with baseline models across three mobile manipulation tasks in various simulated 3D environments. In \cref{sec:real_exp}, we validate our method in real-world 3D household settings. Finally, \cref{sec:limits_and_future_work} discusses the limitations and outlines potential directions for future research, followed by \cref{sec:conclusion} where we conclude the paper. 

\begin{figure*}[t!]
    \centering
    \includegraphics[width=\linewidth]{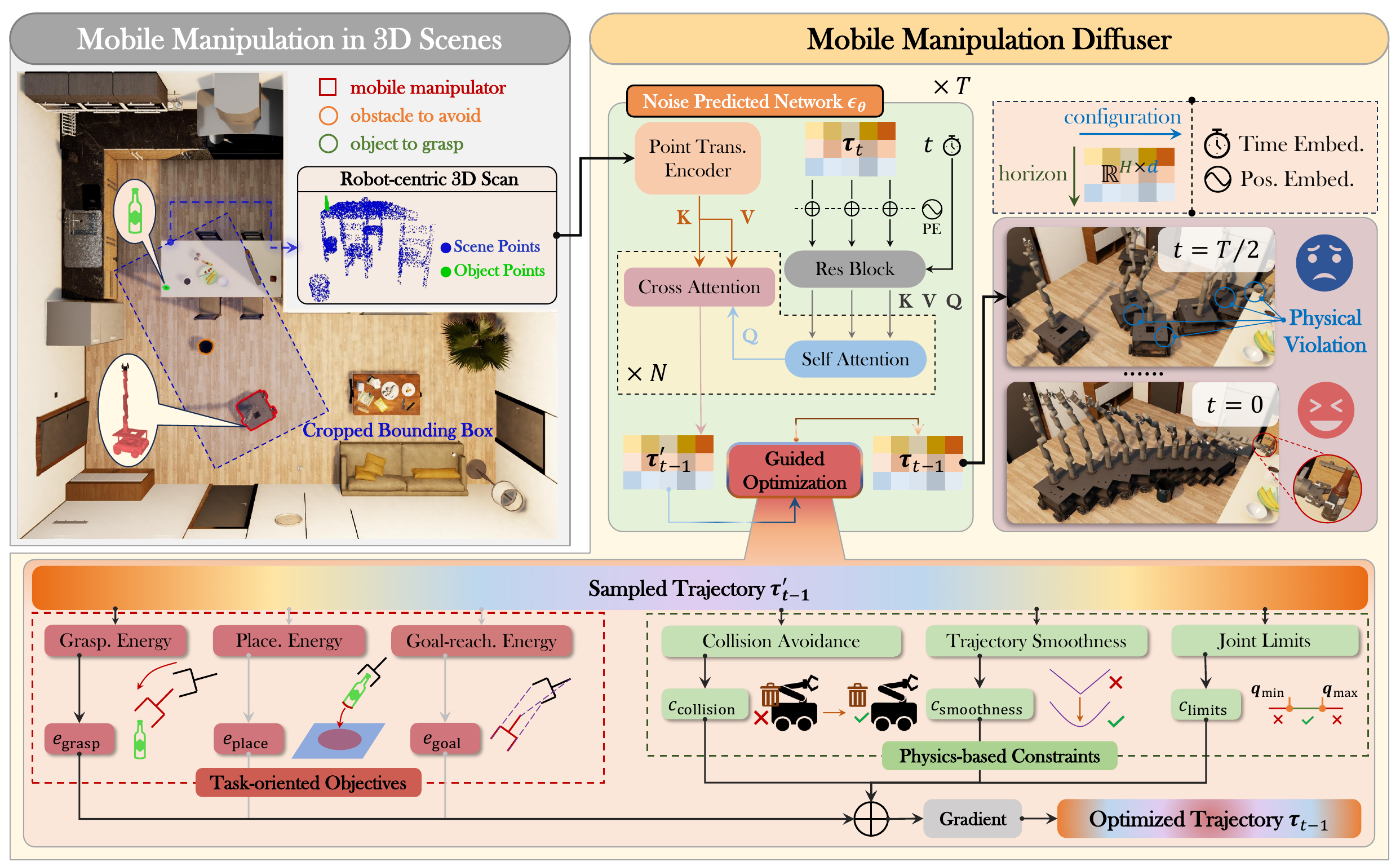}
    \caption{Overview of the \model, a diffusion-based motion planner designed to sample and optimize whole-body coordinated trajectories directly from natural 3D scans, efficacious for mobile manipulation in 3D scenes. Using robot-centric 3D scans as visual input, \model employs an iterative denoising process to generate task-specific trajectories. It optimizes the sampled results at each denoising diffusion step guided by cost and energy functions, ensuring physical plausibility and task completion of generated trajectories.}
    \label{fig:overview}
\end{figure*}

\section{Mobile Manipulation Diffuser} \label{sec:m2diffuser}
\subsection{Problem Statement and Diffusion Model}
Given the robot-centric 3D scan of a scene $\mathcal{S}$, \model aims to generate an efficient and coordinated trajectory that fulfills the task objective $\mathcal{O}$, enabling the robot to complete tasks such as object grasping, object placement, or reaching a target pose without physical violations. We denote the trajectory as $\boldsymbol{\tau}=\left(\mathbf{q}_0, \cdots, \mathbf{q}_i, \cdots, \mathbf{q}_H\right)$, where $\mathbf{q}_i \in \mathbb{R}^{d}$ is the robot's joint position and the trajectory is discreted by the task horizon $H$. Here we assume that a low-level controller can robustly drive the robot's configuration $\mathbf{q}_i$ to $\mathbf{q}_{i+1}$ as long as they are physically feasible.

Diffusion model~\cite{ho2020denoising} consists of a forward and a reverse diffusion process, respectively corresponding to the model training and inference. During the forward process $q\left(\boldsymbol{\tau}_t \mid \boldsymbol{\tau}_{t-1}\right)$, the initial data $\boldsymbol{\tau}_0 \sim q\left(\boldsymbol{\tau}_0\right)$ sampled from the dataset is perturbed by adding gradually-decreased noise, which eventually turns the data into Gaussian noise $\boldsymbol{\tau}_{T}$. In the reverse process, the data is reconstructed from $\boldsymbol{\tau}_{T}$ following an iterative denoising process with learned Gaussian kernels.     
\model formulates mobile manipulation as trajectory optimization and solves it with the spirit of optimization as inference, \ie, by sampling the trajectory-level distribution learned by the diffusion model. Leveraging the diffusion model with loss-guided sampling and flexible conditioning, \model models the probability of mobile manipulation trajectory conditioned on 3D scan $\mathcal{S}$ and objective $\mathcal{O}$ as:
\begin{equation}
    \label{eq:traj_with_obj}
    \begin{aligned}
    p\left(\boldsymbol{\tau}_0 \mid \mathcal{S}, \mathcal{O}\right)=\int p\left(\boldsymbol{\tau}_T \mid \mathcal{S},\mathcal{O}\right) \prod_{t=1}^T p\left(\boldsymbol{\tau}_{t-1} \mid \boldsymbol{\tau}_t, \mathcal{S}, \mathcal{O}\right) \mathrm{d} \boldsymbol{\tau}_{1: T},
    \end{aligned}
\end{equation}
where $T$ denotes the maximum time step in diffusion process, and $p\left(\boldsymbol{\tau}_T \mid \mathcal{S},\mathcal{O}\right)$ is a standard Gaussian distribution. To sample from $p\left(\boldsymbol{\tau}_0 \mid \mathcal{S}, \mathcal{O}\right)$, we must iteratively sample from the conditional distribution $p\left(\boldsymbol{\tau}_{t-1} \mid \boldsymbol{\tau}_t, \mathcal{S}, \mathcal{O}\right)$, which follows
\begin{equation}
    \label{eq:p_tau_t_s_o}
    \begin{aligned}
    p\left(\boldsymbol{\tau}_{t-1} \mid \boldsymbol{\tau}_t, \mathcal{S}, \mathcal{O}\right)=&\frac{p_\theta\left(\boldsymbol{\tau}_{t-1} \mid \boldsymbol{\tau}_t, \mathcal{S}\right) p_\phi\left(\mathcal{O} \mid \boldsymbol{\tau}_{t-1}, \mathcal{S}\right)}{p(\mathcal{O} \mid \mathcal{S})} \\ 
    \propto &p_{\theta}\left(\boldsymbol{\tau}_{t-1} \mid \boldsymbol{\tau}_t, \mathcal{S}\right) p_{\phi}\left(\mathcal{O} \mid \boldsymbol{\tau}_{t-1}, \mathcal{S}\right).
    \end{aligned}
\end{equation}

\subsection{Trajectory Generation via Conditional Diffusion} \label{sec:mogen_diffusion}  
$p_{\theta}\left(\boldsymbol{\tau}_{t-1} \mid \boldsymbol{\tau}_t, t, \mathcal{S}\right)$ represents the probability of generating scene-conditioned trajectory $\boldsymbol{\tau}_{t-1}$ at denoising step $t$ and is independent of task objective $\mathcal{O}$. In this work, we model it using a scene-conditioned diffusion model similar to~\cite{huang2023diffusion}. According to the formulation of the diffusion model in~\cite{ho2020denoising}, it can be written as
\begin{equation}
    \label{eq:conditional_diffusion}
    \begin{aligned}
    p_\theta\left(\boldsymbol{\tau}_{t-1} \mid \boldsymbol{\tau}_t, \mathcal{S}\right)=\mathcal{N}\left(\boldsymbol{\tau}_{t-1} ; \boldsymbol{\mu}_\theta\left(\boldsymbol{\tau}_t, t, \mathcal{S}\right), \boldsymbol{\Sigma}_\theta\left(\boldsymbol{\tau}_t, t, \mathcal{S}\right)\right).
    \end{aligned}
\end{equation}

For simplicity, we only learn the mean $\boldsymbol{\mu}_{\theta}$, while the covariance $\boldsymbol{\Sigma}_{\theta}$ is decided by noise schedules. Ingeniously,~\cite{ho2020denoising} formulates $\boldsymbol{\mu}_{\theta}$ as
\begin{equation}
    \label{eq:mu_theta_and_noise}
    \begin{aligned}
    \boldsymbol{\mu}_{\theta}\left(\boldsymbol{\tau}_t, t, \mathcal{S}\right)=\frac{1}{\sqrt{\alpha_t}}\left(\boldsymbol{\tau}_t-\frac{1-\alpha_t}{\sqrt{1-\bar{\alpha}_t}} \boldsymbol{\epsilon}_{\boldsymbol{\theta}}\left(\boldsymbol{\tau}_t, t, \mathcal{S}\right)\right),
    \end{aligned}
\end{equation}
where ${\alpha}_t$ and $\bar{\alpha}_t$ are defined by noise schedules in the forward process~\cite{ho2020denoising, song2021score, song2019generative}. Then $\boldsymbol{\mu}_{\theta}$ can be learned with a noise prediction network $\boldsymbol{\epsilon}_{\theta}$ via a MSE loss:
\begin{equation}
    \label{eq:loss_function}
    \begin{aligned}
    \mathcal{L}_\theta\left(\boldsymbol{\tau}_0 \mid \mathcal{S}\right) & =\mathbb{E}_{{t}, \boldsymbol{\epsilon}, \tau_0}\left[\left\|\boldsymbol{\epsilon}-\boldsymbol{\epsilon}_\theta\left(\sqrt{\bar{\alpha}_t} \boldsymbol{\tau}_0+\sqrt{1-\bar{\alpha}_t} \boldsymbol{\epsilon}, t, \mathcal{S}\right)\right\|^2\right] \\
    & =\mathbb{E}_{{t}, \boldsymbol{\epsilon}, \tau_0}\left[\left\|\boldsymbol{\epsilon}-\boldsymbol{\epsilon}_\theta\left(\boldsymbol{\tau}_t, t, \mathcal{S}\right)\right\|^2\right],
    \end{aligned}
\end{equation}
with $t \sim \mathcal{U}(1, T)$, $\boldsymbol{\epsilon} \sim \mathcal{N}(\mathbf{0}, \mathbf{I})$ and $\tau_0 \sim q\left(\tau_0\right)$. Specifically, we utilize the architecture $\boldsymbol{\epsilon}_\theta$ (as shown in \cref{fig:overview}) to predict the noise at each diffusion step.

\subsection{Trajectory Optimization via Guided Sampling}
\label{sec:trajopt}

$p_{\phi}\left(\mathcal{O} \mid \boldsymbol{\tau}_{t-1}, \mathcal{S}\right)$ indicates the likelihood of accomplishing the task objective $\mathcal{O}$ in scene $\mathcal{S}$ with trajectory $\boldsymbol{\tau}_{t-1}$. Practically, achieving $\mathcal{O}$ also implies that the trajectory $\boldsymbol{\tau}_{t-1}$ is subject to the constraints imposed by the 3D scene $\mathcal{S}$ and robot embodiment. Therefore, we write $p_{\phi}\left(\mathcal{O} \mid \boldsymbol{\tau}_{t-1}, \mathcal{S}\right)$ in its exponential and decompose it \wrt the task objective and constraints:
\begin{equation}
    \label{eq:likelihood_task}
    \begin{aligned}
    p_{\phi}\left(\mathcal{O} \mid \boldsymbol{\tau}_{t-1}, \mathcal{S}\right) &\propto \exp \left(\varphi\left(\boldsymbol{\tau}_{t-1}, \mathcal{S}\right)\right) \\
    &=\exp \left(- e\left(\boldsymbol{\tau},\mathcal{S}\right) - \sum_i \lambda_i c_i\left(\boldsymbol{\tau},\mathcal{S}\right) \right),
    \end{aligned}
\end{equation}
where $e\left(\boldsymbol{\tau},\mathcal{S}\right)$ represents the energy function for task completion, and each $c_i\left(\boldsymbol{\tau},\mathcal{S}\right)$ represents the cost function of violating a physical constraint. The energy function varies in different tasks, such as object grasping, object placement, or reaching a goal configuration. We define multiple cost functions that penalize failures to meet collision avoidance, trajectory smoothness, and joint limit constraints, balanced with weight $\lambda_i$. The design of these functions is detailed in \cref{sec:energy_functions} and \cref{sec:cost_functions}.

According to the definition of $\boldsymbol{\Sigma}_{t}$ in~\cite{ho2020denoising}, as the diffusion time step $t$ approaches 0, the noise covariance $\left\|\boldsymbol{\Sigma}_{t}\right\|\rightarrow0$. Consequently, we can approximate $\log p_{\phi}\left(\mathcal{O} \mid \boldsymbol{\tau}_{t-1}, \mathcal{S}\right)$ using a first-order Taylor expansion around $\boldsymbol{\tau}_{t-1}=\boldsymbol{\mu}_t$ following~\cite{carvalho2023motion,huang2023diffusion}:
\begin{equation}
    \label{eq:log_likelihood_traj_opt}
    \begin{aligned}
    \log p_{\phi}\left(\mathcal{O} \mid \boldsymbol{\tau}_{t-1}, \mathcal{S}\right) \approx \left(\boldsymbol{\tau}_{t-1}-\boldsymbol{\mu}_t\right)^T \boldsymbol{g} + C,
    \end{aligned}
\end{equation}
where $\boldsymbol{\mu}_t=\boldsymbol{\mu}_{\theta}\left(\boldsymbol{\tau}_t, t, \mathcal{S}\right)$, $\boldsymbol{\Sigma}_t=\boldsymbol{\Sigma}_\theta\left(\boldsymbol{\tau}_t, t, \mathcal{S}\right)$, $C$ is a constant, and the gradient
\begin{equation}
    \label{eq:loss_guidance}
    \begin{aligned}
    \boldsymbol{g}&=\nabla_{\boldsymbol{\tau}_{t-1}} \log p_{\phi}\left(\mathcal{O} \mid \boldsymbol{\tau}_{t-1}, \mathcal{S}\right)|_{\boldsymbol{\tau}_{t-1}=\boldsymbol{\mu}_t} \\
    &=\nabla_{\boldsymbol{\tau}_{t-1}} \varphi\left(\boldsymbol{\tau}_{t-1}, \mathcal{S}\right)|_{\boldsymbol{\tau}_{t-1}=\boldsymbol{\mu}_t}.
    \end{aligned}
\end{equation}
Further, we can rewrite \cref{eq:p_tau_t_s_o} as:
\begin{equation}
    \label{eq:loss_guided_diffusion}
    \begin{aligned}
    p\left(\boldsymbol{\tau}_{t-1} \mid \boldsymbol{\tau}_t, \mathcal{S}, \mathcal{O}\right)=\mathcal{N}\left(\boldsymbol{\tau}_{t-1}; \boldsymbol{\mu}_t+\boldsymbol{\Sigma}_t \boldsymbol{g}, \boldsymbol{\Sigma}_t\right),
    \end{aligned}
\end{equation}
which follows a Gaussian distribution that is easy to sample from. Then trajectory optimization with \model is to iteratively apply guided sampling until convergence. We present the complete inference process of \model in \cref{alg:mpidiffuser-inference} and the training process in \cref{alg:mpidiffuser-training}.

\begin{figure}[t!]
    \centering
    \includegraphics[width=\linewidth]{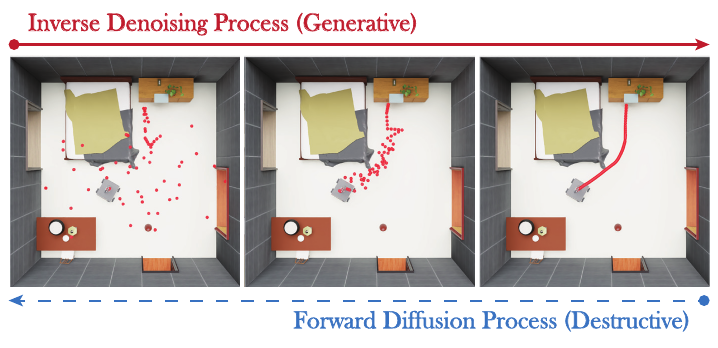}
    \caption{The diffusion and denoising process of \model. The example shows the diffusion and denoising process of the robot's end effector trajectory in a grasping task (\eg, grasping a book).}
    \label{fig:diffusion_process}
\end{figure}

\subsection{Defining Task Objective as Energy Function}\label{sec:energy_functions}
As described in \cref{sec:trajopt}, trajectory optimization with \model relies on defining the task objective with an energy function. We consider three alternative task objectives, \ie, object grasping, object placement, and goal reaching. Below, we elaborate the definition of the corresponding energy functions.

\paragraph{Grasping Energy} For a 3D object, it's usually nontrivial to define an energy function to measure the quality of a $\mathrm{SE}(3)$ grasping pose due to the multi-modal natural of the solution space. Therefore, we define a data-driven grasping energy function following~\cite{urain2023se} to guide the diffusion process to jointly optimize grasp sampling and trajectory generation:
\begin{equation}
    \label{eq:grasp_energy}
    \begin{aligned}
    e_{\text{grasp}}={E}_{\theta}\left(P_{o}, \phi_{ee}^{\mathbf{H}}\left(\mathbf{q}_{H-1}\right), t\right),
    \end{aligned}
\end{equation}
where ${E}_{\theta}$ is pre-trained as in~\cite{urain2023se}, $P_{o}$ is the point cloud of target object, $\phi_{ee}^{\mathbf{H}}(\cdot): \mathbb{R}^{d} \rightarrow \mathrm{SE}(3)$ is the robot's forward kinematics that maps joint position into end effector pose, and $\mathbf{q}_{H-1}$ is the last joint position of sampled trajectories and $t$ is current diffusion time step. 

\begin{algorithm}[t!]
    \fontsize{9}{9}\selectfont{}
    \caption{Training of \model}
    \label{alg:mpidiffuser-training}
    \SetKwInOut{module}{Modules}
    \KwIn{Trajectories in 3D scene $({\boldsymbol\tau}_0, \mathcal{S})$, Noise prediction model of conditional diffusion $\boldsymbol{\epsilon}_{\theta}$, learning rate $\eta$ and noise schedule terms $\bar{\alpha}_t$} 
    
    \nonl \textcolor{blue}{\texttt{// train base generation model}} \\
    \Repeat{\text{converged}}{  
        \nonl \textcolor{blue}{\texttt{// sample trajectory}} \\
        ${\boldsymbol\tau}_0 \sim p({\boldsymbol\tau}_0 \mid \mathcal{S})$ \\

        \nonl \textcolor{blue}{\texttt{// sample noise and iteration step}} \\
        $\boldsymbol{\epsilon} \sim \mathcal{N}({\bf 0}, {\bf I})$, $t \sim \mathcal{U}(1, T)$ \\
        
        \nonl \textcolor{blue}{\texttt{// compute loss and update gradient}} \\
        ${\boldsymbol\tau}_t = \sqrt{\bar{\alpha}_t}{\boldsymbol\tau}_0 + \sqrt{1 - \bar{\alpha}_t}\boldsymbol{\epsilon}$ \\
        \vspace{0.1em} 
        $\theta = \theta - \eta\nabla_\theta \|\boldsymbol{\epsilon} - \boldsymbol{\epsilon}_{\theta}({\boldsymbol\tau}_t, t, \mathcal{S})\|^2$
    }  
\end{algorithm}

\begin{algorithm}[t!]
    \fontsize{9}{9}\selectfont{}
    \caption{Inference of \model}
    \label{alg:mpidiffuser-inference}
    \SetKwInOut{module}{Modules}
    \SetKwProg{Fn}{function}{:}{}
    \SetKwFunction{sample}{{\bf sample}}
    \module{Noise prediction model of conditional diffusion $\boldsymbol{\epsilon}_{\theta}$, initial joint position $\mathbf{q}_0$, energy function $e\left(\cdot\right)$, cost functions $\{c_i\left(\cdot\right)\}$ with weights $\{\lambda_i\}$} 

    \nonl \textcolor{blue}{\texttt{// one-step guided sampling}} \\
    \Fn{\sample$(\boldsymbol{\tau}_{t},\varphi)$}{
        \nonl \textcolor{blue}{\texttt{// compute the mean and covariance}} \\
        $\boldsymbol{\mu}_t=\boldsymbol{\mu}_{\boldsymbol{\theta}}\left(\boldsymbol{\tau}_t, t, \mathcal{S}\right), \boldsymbol{\Sigma}_t=\boldsymbol{\Sigma}_\theta\left(\boldsymbol{\tau}_t, t, \mathcal{S}\right)$ \\
        \nonl \textcolor{blue}{\texttt{// compute gradient}} \\
        $\boldsymbol{g}=\nabla_{\boldsymbol{\tau}_{t-1}} \varphi\left(\boldsymbol{\tau}_{t-1}, \mathcal{S}\right)|_{\boldsymbol{\tau}_{t-1}=\boldsymbol{\mu}_t}$ \\
        \nonl \textcolor{blue}{\texttt{// sample with guidance}} \\
        $\boldsymbol\tau_{t-1} \sim \mathcal{N}\left(\boldsymbol{\tau}_{t-1}; \boldsymbol{\mu}_t+\boldsymbol{\Sigma}_t \boldsymbol{g}, \boldsymbol{\Sigma}_t\right)$ \\
        \nonl \textcolor{blue}{\texttt{// set initial state}} \\
        $\boldsymbol{\tau}_{t-1}[0]=\mathbf{q}_0$ \\
        \Return $\boldsymbol\tau_{t-1}$ \\
    } 
    \vspace{1em} 
    
    \nonl \textcolor{blue}{\texttt{// trajectory optimization}} \\
    \KwIn{initial trajectory $\boldsymbol\tau_{T} \sim \mathcal{N}({\bf 0}, {\bf I}),{\tau}_{T}[0]:=\mathbf{q}_0$}
    \nonl \textcolor{blue}{\texttt{// iterative denoising by guided sampling}} \\
    \For{$t=T,\cdots,0$}{       
        $\boldsymbol{\tau}_{t-1}=\sample(\boldsymbol{\tau}_{t}, - \left[e + \sum_i \lambda_i c_i\right])$\\
    }
    \nonl \textcolor{blue}{\texttt{// additional steps to improve convergence}} \\
    \For{$k=K,\cdots,1$}{      
        $\boldsymbol{\tau}_{0}=\sample(\boldsymbol{\tau}_{1},-\left[e + \sum_i \lambda_i c_i\right])$\\
        $\boldsymbol{\tau}_{1}=\boldsymbol{\tau}_{0}$\\
    }
    \Return $\boldsymbol\tau_0$ \\
\end{algorithm}

\paragraph{Placement Energy} We define a placement energy function to guide the robot to place the object on the target area with physical plausibility:
\begin{equation}
    \label{eq:place_energy}
    \begin{aligned}
    e_{\text{place}}=&\sum_{p_{1i} \in P_1} \min _{p_{2 j} \in P_2}\left(\left\|p_{1 i}-p_{2 j}\right\|_2^2\right)+ \\
    &\sum_{p_{2 j} \in P_2} \min _{p_{1 i} \in P_1}\left(\left\|p_{2 j}-p_{1 i}\right\|_2^2\right),
    \end{aligned}
\end{equation}
where $P_2$ denotes point cloud of the given target area (\eg, desk surface), and $P_1$ denotes point cloud of the object's placement surface that transforms with $\mathbf{q}_{H-1}$ and the grasping pose. Notably, $P_2$ can be predicted by O2O-Afford~\cite{mo2022o2o} or specified manually, while $P_1$ can be detected by UOP-Net~\cite{noh2024learning}. In practice, we obtain these point clouds ahead of time in a pre-processing step before we solve for the trajectory.

\paragraph{Goal-reaching Energy} The goal-reaching task~\cite{fishman2022mpinets} is to reach an end effector pose represented as the rendered end effector point cloud. We define its energy function to punish the chamfer distance between the goal point cloud and the actual end effector point cloud at configuration $\mathbf{q}_{H-1}$:
\begin{equation}
    \label{eq:goal_reaching_energy}
    \begin{aligned}
    e_{\text{goal}}=&\sum_{{p}_{ee}^i \in {P}_{ee}} \min _{p_{g}^j \in {P}_{g}}\left(\left\|{p}_{ee}^i-p_{g}^j\right\|_2^2\right)+ \\
    &\sum_{p_{g}^i \in {P}_{g}} \min _{{p}_{ee}^j \in {P}_{ee}}\left(\left\|p_{g}^i-{p}_{ee}^j\right\|_2^2\right),
    \end{aligned}
\end{equation}
where ${P}_{g}$ denotes the goal point cloud, and ${P}_{ee}$ denotes the actual point cloud of end effector at $\mathbf{q}_{H-1}$.


\subsection{Defining Physical Constraints as Cost Functions}\label{sec:cost_functions}

In the following, we further define the list of cost functions that penalize the violation of physical constraints. 

\paragraph{Collision-avoidance Cost} We define the collision-avoidance cost function to penalize the physical collision between the scene and the robot. Instead of calculating mesh collision between scene objects and robots, we estimate the collision depth between \ac{sdf} of the scene and $N$ sampled points on robot's surface. Then the collision-avoidance cost is defined following~\cite{ratliff2009chomp}:
\begin{equation}
    \label{eq:collision_cost}
    \begin{aligned}
    c_{\text{collision}}=\sum_i\sum_h\Phi_s\left({p}^i_h\right), 
    \end{aligned}
\end{equation}
where
\begin{equation}
    \label{eq:sdf_function}
    \begin{aligned}
    \Phi_s\left({p}_h^i\right)=\left\{\begin{array}{cl}
    -\mathcal{D}_s\left({p}_h^i\right)+\frac{1}{2} \varepsilon_c & \text {if } \mathcal{D}_s\left({p}_h^i\right)<0, \\
    \frac{1}{2 \varepsilon_c}\left(\mathcal{D}_s\left({p}_h^i\right)-\varepsilon\right)^2 & \text {if } 0 \leqslant \mathcal{D}_s\left({p}_h^i\right) \leqslant \varepsilon_c, \\
    0 & \text {otherwise.}\end{array}\right.
    \end{aligned}
\end{equation}
We denote a surface point on the robot's end effector at time step $h$ as ${p}^i_h$, and its signed distance in the 3D scene as $\mathcal{D}_s({p}^i_h)$. We set a safety margin $\varepsilon_c > 0$ for collision avoidance. 


\paragraph{Trajectory Smoothness Cost} The smoothness of robot trajectories is essential to prevent abrupt changes in speed and acceleration and improve safety. We define the trajectory smoothness cost function to minimize the difference in joint velocity between adjacent time steps:
\begin{equation}
    \label{eq:traj_smooth_cost}
    \begin{aligned}
    c_{\text{smoothness}}=\sum_i\sum_h\left\|{p}^{i+2}_h-2{p}^{i+1}_h+{p}^i_h\right\|^2_2.
    \end{aligned}
\end{equation}

\paragraph{Joint Limit Cost} We define the joint limit cost function to punish the violation of joint limits $\mathbf{q}_{\max}$ and $\mathbf{q}_{\min}$:
\begin{equation}
    \label{eq:joint_limits_cost}
    \begin{aligned}
    c_{\text{limit}}=\sum_h\mathcal{J}\left(\mathbf{q}_h\right), 
    \end{aligned}
\end{equation}
where,
\begin{equation}
    \label{eq:joint_limits_function}
    \begin{aligned}
    \mathcal{J}\left(\mathbf{q}_h\right)= \left\{\begin{array}{cl}\left\|\mathbf{q}_{\text{lower}}-\mathbf{q}_h\right\|_2^2 & \text {if } \mathbf{q}_h<\mathbf{q}_{\text{lower}}, \\ 0 & \text {if } \mathbf{q}_{\text{lower}} \leq \mathbf{q}_h \leq \mathbf{q}_{\text{upper}}, \\ \left\|\mathbf{q}_{\text{upper}}-\mathbf{q}_h\right\|_2^2 & \text {otherwise.}\end{array}\right.
    \end{aligned}
\end{equation}
Here, $\mathbf{q}_{\text{lower}}=\mathbf{q}_{\min}+\varepsilon_l$, $\mathbf{q}_{\text{upper}}=\mathbf{q}_{\max}-\varepsilon_l$, $\mathbf{q}_h$ denotes the joint position at the time step $h$, and $\varepsilon_l > 0$ defines the safety margin for joint limit violation.

\subsection{Model Architecture} 
As shown in \cref{fig:overview}, \model expects three inputs, the current diffusion time step $t$, intermediate sampled trajectory $\boldsymbol{\tau}_{t}$ and a scene point cloud cropped from the 3D scan based on the bounding box around the robot. The noise prediction network $\boldsymbol{\epsilon}_{\theta}$ builds on previous work~\cite{huang2023diffusion} and adopts a PointTransformer to encode the 3D observation and output latent per-point features as the key and value for the cross-attention module. Moreover, $\boldsymbol{\epsilon}_{\theta}$ utilizes a fully-connected layer along with positional embedding to extract high-dimensional features from the trajectory. These features are then fused with the diffusion time step embedding through a ResBlock. The fused results are subsequently fed into a self-attention module and served as the query for the cross-attention module. Following that, $\boldsymbol{\epsilon}_{\theta}$ estimates the noise in the current time step by employing a feedforward layer. Finally, leveraging the estimated noise, \model samples the next intermediate trajectory $\boldsymbol{\tau}_{t-1}$ with guidance from the energy and cost functions.

\begin{table*}[ht!]
    \centering
    \caption{Statistical analysis of whole-body trajectory datasets.}
    \resizebox{\textwidth}{!}{%
        \setlength{\tabcolsep}{3pt}
        \setstretch{1.5}
        \begin{tabular}{cc ccccccccccccccccc}
            \toprule
            &  \multirow{1}[2]{*}{\makecell{\makebox[2.5cm][c]{Trajectory Dataset}}} & \includegraphics[width=0.035\textwidth]{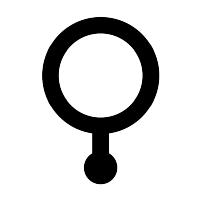} & 
            \includegraphics[width=0.032\textwidth]{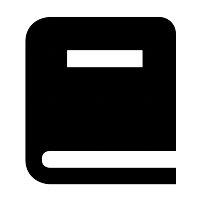} & \includegraphics[width=0.033\textwidth]{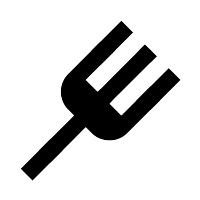} & \includegraphics[width=0.026\textwidth]{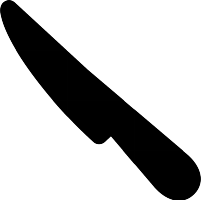} & \includegraphics[width=0.028\textwidth]{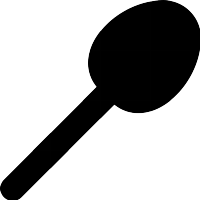} & \includegraphics[width=0.030\textwidth]{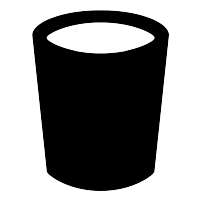} & \includegraphics[width=0.033\textwidth]{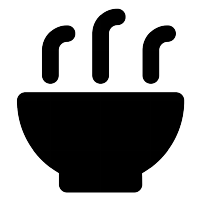} & \includegraphics[width=0.033\textwidth]{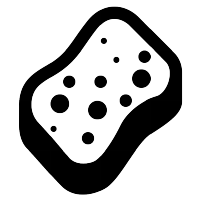} & \includegraphics[width=0.042\textwidth]{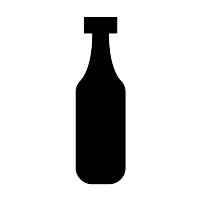} & \includegraphics[width=0.035\textwidth]{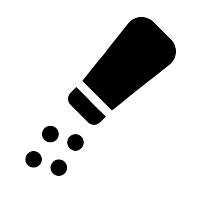} & \includegraphics[width=0.033\textwidth]{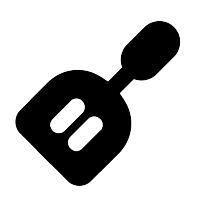} & \includegraphics[width=0.030\textwidth]{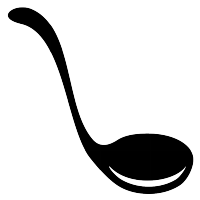} & \includegraphics[width=0.030\textwidth]{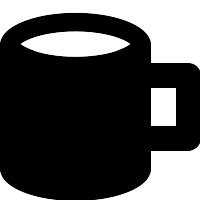} & \includegraphics[width=0.023\textwidth]{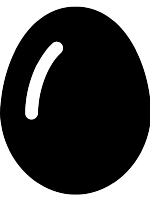} & \includegraphics[width=0.031\textwidth]{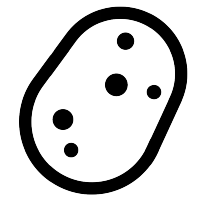} & \includegraphics[width=0.037\textwidth]{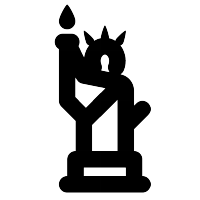} & \includegraphics[width=0.035\textwidth]{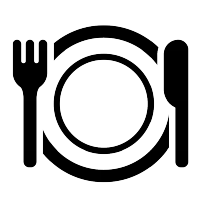}\\
            & & Pan & Book & Fork & Knife & Spoon & Cup & Bowl & Sponge & Bottle & Shaker & Spatula & Ladle & Mug & Egg & Potato & Statue & Plate\\
            
            \midrule
            
            \rowcolor{MistyRose}
            & {Grasp. Set One} & $771/84$ & $548/71$ & $275/31$ & $341/34$ & $751/80$ & $373/33$ & $229/21$ & $592/78$ & $1015/129$ & $598/77$ & $360/49$ & $33/4$ & $360/91$ & $116/16$ & $119/8$ & $773/79$ & $1324/129$\\
            \rowcolor{MistyRose}
            & {Grasp. Set Two} & $0/21$ & $0/17$ & $0/13$ & $0/21$ & $0/25$ & $0/34$ & $0/26$ & $0/21$ & $0/26$ & $0/21$ & $0/21$ & $0/21$ & $0/21$ & $0/21$ & $0/26$ & \ding{55} & \ding{55}\\

            \midrule
            \midrule
            
            \rowcolor{LightCyan1}
            & {Place. Set One} & $407/44$ & $626/77$ & $278/26$ & $615/59$ & $611/68$ & $140/13$ & $187/25$ & $967/111$ & $1084/105$ & $1084/105$ & $159/17$ & $1352/144$ & $364/48$ & $19/0$ & $12/3$ & $16/0$ & $184/25$\\
            \rowcolor{LightCyan1}
            & {Place. Set Two} & \ding{55} & $0/48$ & $0/47$ & \ding{55} & $0/48$ & $0/48$ & $0/6$ & \ding{55} & $0/24$ & $0/1$ & $0/48$ & $0/24$ & $0/8$ & $0/3$ & \ding{55} & \ding{55} & \ding{55}\\
 
            \bottomrule
        \end{tabular}%
        
    }%
    \begin{tablenotes}
        \footnotesize
        \item The \textcolor{grasping}{grasping} and \textcolor{placement}{placement} trajectories respectively collected from 24 and 32 simulated scenes are divided into two subsets: set one and set two (generalization evaluation set). The data in the table indicates the number of trajectories used for training (left) and testing (right). We train our model and baselines only on set one and respectively evaluate models performance on both set one (seen scenes) and set two (unseen scenes).
    \end{tablenotes}
    \label{table:traj_data}
\end{table*}

\begin{figure}[t!]
    \centering
    \includegraphics[width=\linewidth]{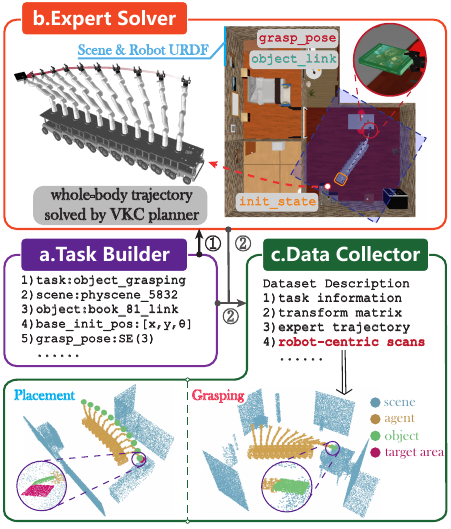}
    \caption{Dataset collection procedure. (a) The \textbf{Task Builder} enables the construction of mobile manipulation tasks through high-level configurations, including scene and robot URDF, manipulated object link, target end effector goal, and task type. (b) The \textbf{Expert Solver} computes optimal whole-body coordinated trajectories by leveraging the \ac{vkc} algorithm~\cite{jiao2021consolidating,jiao2021efficient}. (c) The \textbf{Data Collector} is responsible for recording the planned trajectories, and processing the segmented point clouds cropped from the perfect 3D scan based on the bounding box around the robot's initial position.}
    \label{fig:dataset_collection}
\end{figure}

\section{Experiments in Simulated 3D Scenes} \label{sec:sim_exp}
\subsection{Dataset Preparation}
To collect a large volume of whole-body coordinated expert trajectories, we utilize the autonomous tool developed in our previous work~\cite{zhang2024m3bench}. The data collection procedure is illustrated in \cref{fig:dataset_collection}.
The collected grasping and placement expert trajectories cover 26 common objects with diverse geometries across 24 and 32 simulated 3D scenes, respectively. Specifically, the 24 scenes used for grasping data collection are divided into two groups. From 17 of these scenes, we collected 10673 grasping expert trajectories, which are split into training and testing sets with a 9:1 ratio. The remaining 335 trajectories, collected from the other 7 scenes, are used exclusively for evaluating the model's generalizability to novel scenes.
Similarly, the 32 scenes used for placement data collection are divided into two groups. From 24 of these scenes, we collect 8996 placement expert trajectories, which are also divided into training and testing sets following the same ratio. The remaining 351 trajectories from the other 8 scenes are used solely for generalizability evaluation. \cref{table:traj_data} displays some of the manipulated objects along with the number of expert demonstrations used for training and testing.
Notably, we reuse the grasping trajectories from our dataset to train and test the target-reaching task.

All our collected expert demonstrations are planned by \ac{vkc} algorithm~\cite{jiao2021consolidating, jiao2021efficient}, an optimization-based global planner specifically designed to solve whole-body trajectory optimization. We use a planning horizon of 50 time steps.
The simulated 3D scenes used in our work are sourced from PhyScene~\cite{yang2024physcene}, a scene synthesis method that generates realistic 3D household scenes with rich interactive objects, tailored for robot learning.

\begin{figure}[t!]
    \centering
    \includegraphics[width=\linewidth]{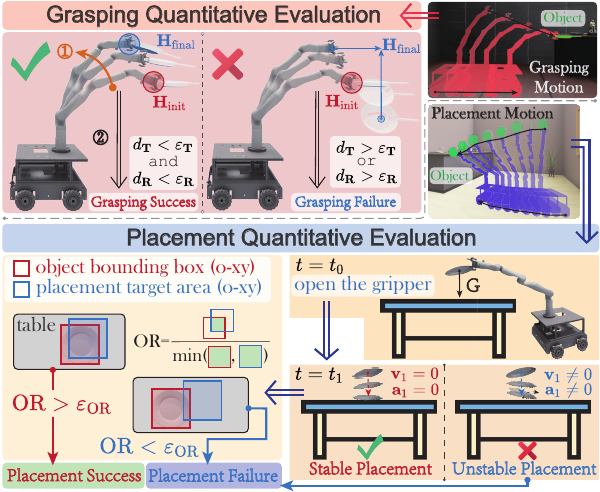}
    \caption{Quantitative evaluation metrics of grasping and placement tasks. Previous work~\cite{batra2020rearrangement, ehsani2021manipulathor, huang2023skill, ni2023towards} evaluate object grasping by the contact between the end effector's bounding sphere and the object surface. This evaluation strategy often fails to reflect how the grasping performs in real-world scenarios. In this paper, we evaluate object grasping and placement quality by success rate in simulated scenes with physical simulation enabled. We use NVIDIA Isaac Sim as the physical simulator.}
    \label{fig:evaluation_sim}
\end{figure}

\begin{figure*}[ht!]
    \centering
    \footnotesize
    \begin{minipage}{\linewidth}
        \centering
        \includegraphics[width=\textwidth]{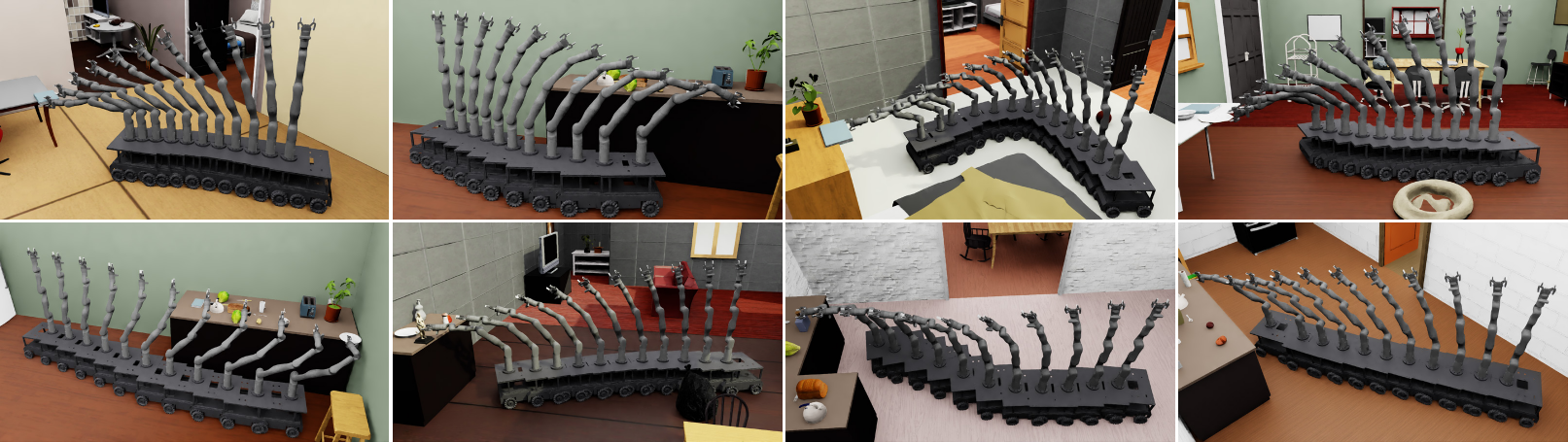} 
        \hfill
        \text{(a) successful grasping trajectories generated by \model} \\[0.5em]
        \label{fig:success_grasping}
    \end{minipage}

    \begin{minipage}{\linewidth}
        \centering
        \includegraphics[width=\textwidth]{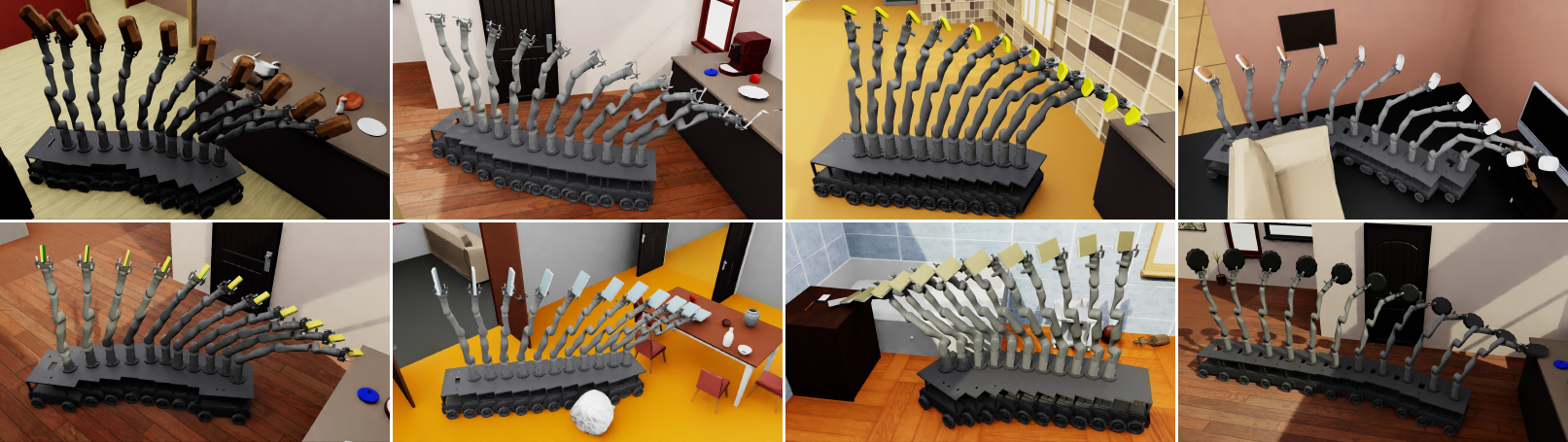} 
        \hfill
        \text{(b) successful placement trajectories generated by \model} 
        \label{fig:success_placement} 
    \end{minipage}

    \caption{Successful trajectories generated by \model on object grasping and placement tasks. These figures illustrate the successful trajectories generated by our method in (a) grasping and (b) placement tasks involving various objects.}
    \label{fig:success_motion_generation}
\end{figure*}

\begin{figure*}[t!]
    \centering
    \footnotesize
    \begin{minipage}{0.32\linewidth}
        \centering
        \includegraphics[width=\linewidth]{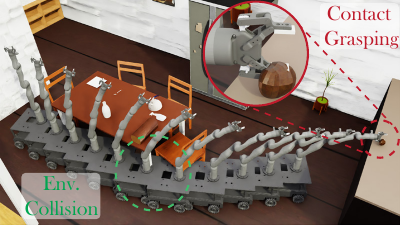} 
        \hfill
        \text{(a) failed object grasping}
        \label{fig:failed_grasping}
    \end{minipage}
    \hfill
    \begin{minipage}{0.32\linewidth}
        \centering
        \includegraphics[width=\linewidth]{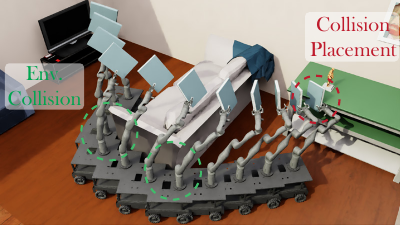}   
        \hfill
        \text{(b) failed object placement}
        \label{fig:failed_placement}
    \end{minipage}
    \hfill
    \begin{minipage}{0.32\linewidth}
        \centering
        \includegraphics[width=\linewidth]{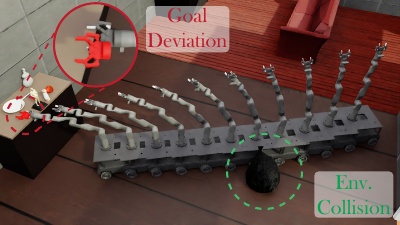}
        \hfill
        \text{(c) failed goal-reaching}
        \label{fig:failed_goal_reach}
    \end{minipage}
    
    \caption{Typical failure cases of baseline models. The trajectories generated by baselines fail to (a) grasp the object due to collisions and physical contact, (b) place the object due to an improper pose, and (c) reach the target end effector goal.}
    \label{fig:failed_motion_generation}
\end{figure*}

\subsection{Mobile Manipulation Tasks for Evaluation} \label{sec:mm_task_eval}
\paragraph{Object Grasping} For a given grasping task, the model input is the segmented and cropped point cloud observed in the robot's initial base frame (see \cref{fig:dataset_collection}). These points encode three segmentation classes: target object $P_{s} \in \mathbb{R}^{4096 \times 3}$, object geometry $P_{o} \in \mathbb{R}^{512 \times 3}$ and optional robot geometry at current state $P_{r} \in \mathbb{R}^{1024 \times 3}$ (only for baselines). Here, the input of \mpiformer is a sequence of these observed point clouds.

We quantitatively define the \textit{successful} grasping with the assistance of NVIDIA Isaac Sim (as shown in \cref{fig:evaluation_sim}). In the last frame of the generated trajectory, we gradually close the gripper to the smallest opening and note the transformation matrix from the object to the end effector as $\mathbf{H}_{\text {init}}=\left[\mathbf{R}_{\text {init}} \mid \mathbf{T}_{\text {init}}\right]$. Then, we lift the arm up a certain height, and the transformation matrix at this point is noted as $\mathbf{H}_{\text {final}}=\left[\mathbf{R}_{\text {final}} \mid \mathbf{T}_{\text {final}}\right]$. If the object is gripped successfully, $d_{\mathbf{T}}=\left\|\mathbf{T}_{\text {init}}-\mathbf{T}_{\text {final}}\right\|_2<\varepsilon_{\mathbf{T}}$ and $d_{\mathbf{R}}=\left\|\operatorname{LogMap}\left(\mathbf{R}_{\text {init}}^{\top} \mathbf{R}_{\text {final}}\right)\right\|<\varepsilon_{\mathbf{R}}$, we consider this grasping successful. In the grasping task evaluation, we set $\varepsilon_{\mathbf{T}}$ to 2cm and $\varepsilon_{\mathbf{R}}$ to 15°.

\paragraph{Object Placement} The input point clouds in placement tasks consist of four types of points (see \cref{fig:dataset_collection}): scene points, points on the object's stable placement surface, points in the target placement area $P_{p} \in \mathbb{R}^{512 \times 3}$, and optional robot surface points at the current state (only for baselines). The target area is defined as the projection of the object's 3D bounding box onto the horizontal plane when stably placed, with an example shown by the blue box in \cref{fig:evaluation_sim}.

We also quantitatively define the \textit{successful} placement in NVIDIA Isaac Sim. In the last frame of the generated motion, we smoothly open the robot's gripper until it's fully open. Assuming that after 600 simulation steps the object no longer moves, and the \ac{or} of the bounding box of the object and the target area in the horizontal direction is above $\varepsilon_{\mathbf{OR}}$, we consider this placement successful (see \cref{fig:evaluation_sim}). In the placement evaluation, $\varepsilon_{\mathbf{OR}}$ is set to 0.5.

\paragraph{Goal-reaching} Given the scene's 3D scan and the surface point cloud of the end effector at the target pose, the motion generator requires to generate a whole-body motion which makes the end effector finally reach the target goal. Input point cloud includes three class points, there are scene points, goal points $P_{g} \in \mathbb{R}^{512 \times 3}$ and optional robot surface points at current state (only for baselines). For simplicity, we reuse the grasping trajectories in the dataset for training and testing of goal-reaching task by replacing object points with the surface points of the end effector at the target pose. If the position and orientation target errors of final end effector are below 4cm and 20° respectively, we consider this generated goal-reaching motion a success.

\begin{table}[t!]
    \centering
    \scriptsize        
    \caption{Performance comparison of different methods on three mobile manipulation tasks.}
    \setlength{\tabcolsep}{3pt}
    \setstretch{1.1}
    \begin{tabular}{cc cc}
        \toprule
         \makecell{\makebox[1.2cm][c]{Test Set}} & \makecell{\makebox[2cm][c]{Methods}} & \makecell{\makebox[2.4cm][c]{Success Rate (\%)$\uparrow$}} & \makecell{\makebox[2.4cm][c]{Solving Time (s)$\downarrow$}} \\
        \midrule
        \rowcolor{MistyRose}
        \cellcolor{white} & \mpinets & $0.00$ & $\backslash$ \\
        \rowcolor{MistyRose}
        \cellcolor{white} & \mpiformer & $3.93$ & $0.51 \pm 0.01$ \\
        \rowcolor{MistyRose}
        \cellcolor{white} & Ours (w/o opt.) & $21.95$ & $\underline{\bm{0.47 \pm 0.16}}$ \\
        \rowcolor{MistyRose}
        \cellcolor{white} & Ours (w/ opt.) & $\underline{\bm{30.54}}$ & $4.74 \pm 0.14$ \\

        \cmidrule{2-4}

        \rowcolor{LightCyan1}
        \cellcolor{white} & \mpinets & $2.33$ & $\underline{\bm{0.63 \pm 0.05}}$ \\
        \rowcolor{LightCyan1}
        \cellcolor{white} & \mpiformer & $0.89$ & $0.64 \pm 0.07$ \\
        \rowcolor{LightCyan1}
        \cellcolor{white} & Ours (w/o opt.) & $4.67$ & $0.77 \pm 0.33$ \\
        \rowcolor{LightCyan1}
        \cellcolor{white} & Ours (w/ opt.) & $\underline{\bm{22.89}}$ & $4.93 \pm 0.64$ \\

        \cmidrule{2-4}
        
        \rowcolor{LightOliveGreen}
        \cellcolor{white} & \mpinets & $3.24$ & $\underline{\bm{0.46 \pm 0.08}}$ \\
        \rowcolor{LightOliveGreen}
        \cellcolor{white} & \mpiformer & $1.18$ & $0.88 \pm 0.01$ \\
        \rowcolor{LightOliveGreen}
        \cellcolor{white} & Ours (w/o opt.) & $25.49$ & $0.55 \pm 0.26$ \\
        \rowcolor{LightOliveGreen}
        \multirow{-12}{*}{
            \makebox[1cm][c]{\rotatebox[origin=c]{90}{
                \cellcolor{white}
                \makebox[2cm][c]{Test Set One (seen scenes)}
        }}} & Ours (w/ opt.) & $\underline{\bm{30.49}}$ & $3.89 \pm 1.16$ \\

        \midrule
        \midrule


        \rowcolor{MistyRose}
        \cellcolor{white} & \mpinets & $0.00$ & $\backslash$ \\
        \rowcolor{MistyRose}
        \cellcolor{white} & \mpiformer & $3.28$ & $0.73 \pm 0.15$ \\
        \rowcolor{MistyRose}
        \cellcolor{white} & Ours (w/o opt.) & $9.25$ & $\underline{\bm{0.66 \pm 0.20}}$ \\
        \rowcolor{MistyRose}
        \cellcolor{white} & Ours (w/ opt.) & $\underline{\bm{14.33}}$ & $6.43 \pm 0.29$ \\

        \cmidrule{2-4}

        \rowcolor{LightCyan1}
        \cellcolor{white} & \mpinets & $0.85$ & $\underline{\bm{0.59 \pm 0.01}}$ \\
        \rowcolor{LightCyan1}
        \cellcolor{white} & \mpiformer & $0.28$ & $0.61 \pm 0.11$ \\
        \rowcolor{LightCyan1}
        \cellcolor{white} & Ours (w/o opt.) & $5.70$ & $0.71 \pm 0.25$ \\
        \rowcolor{LightCyan1}
        \cellcolor{white} & Ours (w/ opt.) & $\underline{\bm{12.25}}$ & $4.23 \pm 0.17$ \\

        \cmidrule{2-4}
        
        \rowcolor{LightOliveGreen}
        \cellcolor{white} & \mpinets & $0.90$ & $\underline{\bm{0.49 \pm 0.12}}$ \\
        \rowcolor{LightOliveGreen}
        \cellcolor{white} & \mpiformer & $0.00$ & $\backslash$ \\
        \rowcolor{LightOliveGreen}
        \cellcolor{white} & Ours (w/o opt.) & $9.19$ & $0.64 \pm 0.21$ \\
        \rowcolor{LightOliveGreen}
        \multirow{-12}{*}{
            \makebox[1cm][c]{\rotatebox[origin=c]{90}{
                \cellcolor{white}
                \makebox[2cm][c]{Test Set Two (unseen scenes)}
        }}} & Ours (w/ opt.) & $\underline{\bm{12.61}}$ & $4.93 \pm 0.25$ \\

        \bottomrule
    \end{tabular}%
     \begin{tablenotes}
        \footnotesize
        \item We qualitatively compare the success rate and solving time of our model with two baseline models across three mobile manipulation tasks (\ie, object \textcolor{grasping}{grasping}, object \textcolor{placement}{placement} and \textcolor{goal_reaching}{goal-reaching}), followed by a systematic evaluation of the performance of three neural motion planners on both familiar training scenes and previously unseen scenes.
    \end{tablenotes}
    \label{table:quantitative_comparison}
\end{table}

\subsection{Experiment Setup}
\paragraph{Baseline Methods} 
To the best of our knowledge, \model is the first attempt to learn a whole-body neural motion planner for achieving the trajectory generation for mobile manipulation in 3D scenes. Reviewing studies akin to our research, we select \mpinets~\cite{fishman2022mpinets} and design \mpiformer as compared baselines.
\begin{itemize}
    \item \mpinets~\cite{fishman2022mpinets} is recognized as the state-of-the-art model for addressing collision-free goal-reaching problem, with demonstrated success in 3D-based tabletop manipulation. \mpinets is a reactive motion planner, generating the entire trajectories based on autoregressive planning. We extend this model to the mobile manipulation domain to simultaneously predict the base-arm coordinated configuration states. To adapt \mpinets for mobile manipulation in 3D environments, we replace the scene-centric observation originally used with the robot-centric observation as the visual input. Since \mpinets represents the environment using simple primitive shapes (\eg, cubes, cylinders), we utilize the method from~\cite{wang2022dual} to convert non-watertight meshes into \ac{sdf}, allowing \mpinets to compute the same collision loss for more complex 3D scenes.
    \item \mpiformer is an advanced variant of the Skill Transformer~\cite{huang2023skill} with three key modifications. First, we integrate the action prediction module and the skill prediction module from the original network into a unified whole-body action generation module. This module directly generates coordinated movements of both the base and arms. Second, we enhance the model by incorporating the visual encoder from \mpinets to process 3D scans, replacing the original depth encoder. Lastly, \mpiformer utilizes the transformer architecture from the Decision Transformer~\cite{chen2021decision} to improve sequence modeling for long-horizon mobile manipulation. 
\end{itemize}

\paragraph{Evaluation Metrics} 
We use some quantitative metrics to evaluate the physical plausibility and task-related completion of generated motion over three diverse tasks.
\begin{itemize}
    \item \textit{Success Rate}: A trajectory is successful if there are no physical violations, and the position and orientation of final end effector completes the specific task. 
    \item \textit{Time}: The wall time of \textit{successful} trajectory generation for solving the specific task.
    \item \textit{Collision Rate}: The rate of self and scene collisions.
    \item \textit{Joint Violation}: The rate of joint values out of the limits.
    \item \textit{Smoothness}: Same as~\cite{fishman2022mpinets}, we compute the \ac{sparc}~\cite{balasubramanian2015analysis} values for joint-space trajectory and end effector trajectory. If all these values are below -1.6, we consider the trajectory to be smooth. The smaller the \ac{sparc} value, the smoother the trajectory.
\end{itemize}

\paragraph{Training Implementation} We implement \model and two baselines in Ubuntu 20.04 with PyTorch, training them on a desktop with an AMD Ryzen 9 7950X 16-Core CPU, two NVIDIA GeForce RTX 4090 GPU, and 128GB of RAM. During the training of \model, we use the Adam optimizer with a learning rate 0.0001 to update the model parameters. The maximum diffusion step is set to 50, and we train the model for 2000 epochs with a batch size of 256 per task from the dataset. 
Specifically, we train \mpinets for 100 epochs using a batch size of 256. The other hyperparameters and configurations for \mpinets remain unchanged from the original setup~\cite{fishman2022mpinets}. 
Likewise, we train \mpiformer for 100 epochs with a batch size of 64, using the AdamW optimizer with an initial learning rate of 0.0008 and a weight decay of 0.003, where a cosine schedule is employed with 10 epochs warmup.
We select the best model in terms of performance on the validation split throughout the training process.
More details about model training can be found in our code.

\begin{figure}[t!]
    \centering
    \footnotesize
    \begin{minipage}{\linewidth}
        \centering
        \includegraphics[width=\textwidth]{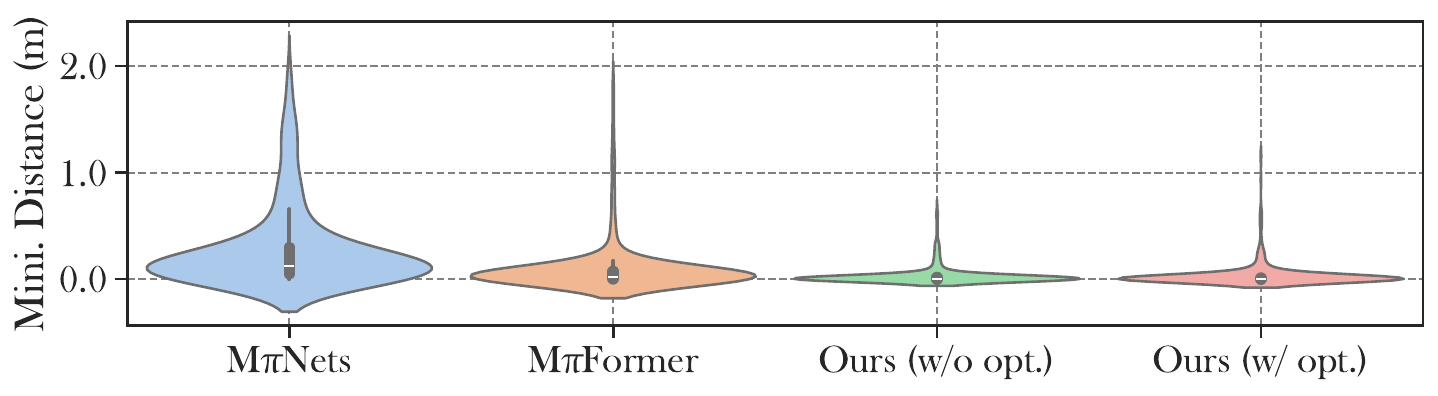} 
        \hfill
        \text{(a) minimum distance between the end effector and the object} \\[0.5em]
        \label{fig:c_dist_all_grasping}
    \end{minipage}
    
    \begin{minipage}{\linewidth}
        \centering
        \includegraphics[width=\textwidth]{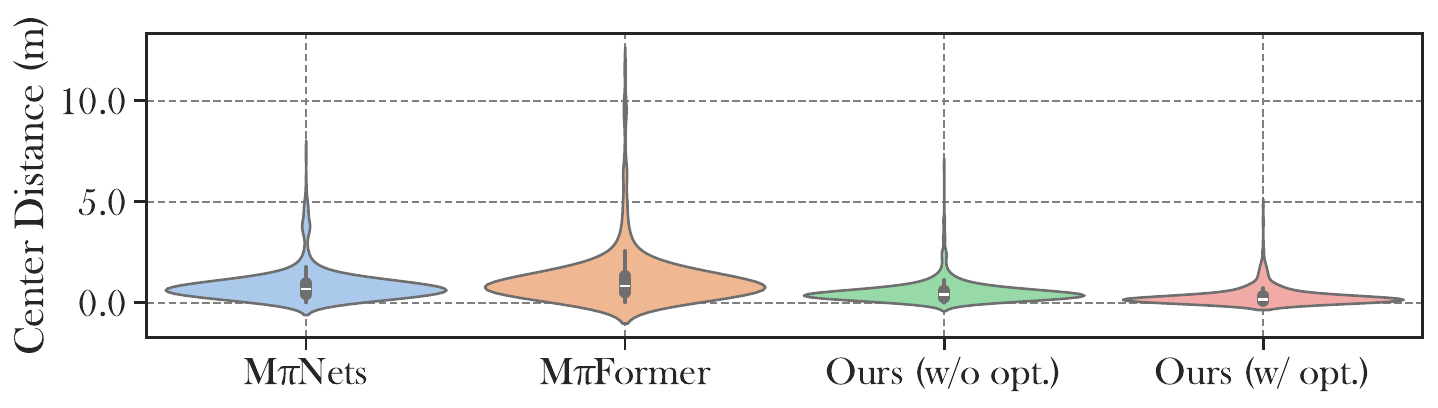} 
        \hfill
        \text{(b) center distance between the object and the target area} \\[0.5em]
        \label{fig:placement_center_dist}
    \end{minipage}

    \begin{minipage}{\linewidth}
        \centering
        \includegraphics[width=\textwidth]{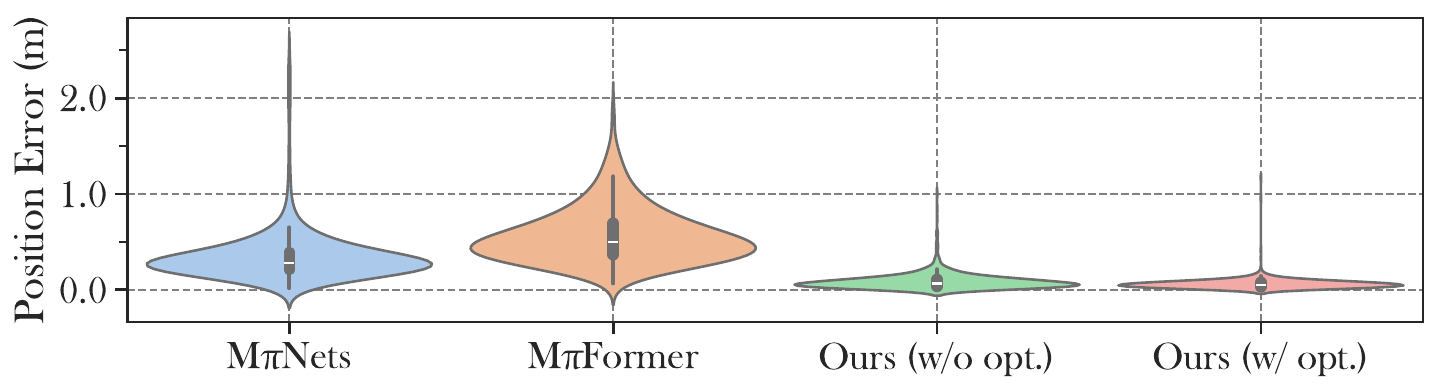} 
        \hfill
        \text{(c) position target error between end effector pose and given pose} \\[0.5em]
        \label{fig:goal_reaching_pos_error}
    \end{minipage}
    
    \begin{minipage}{\linewidth}
        \centering
        \includegraphics[width=\textwidth]{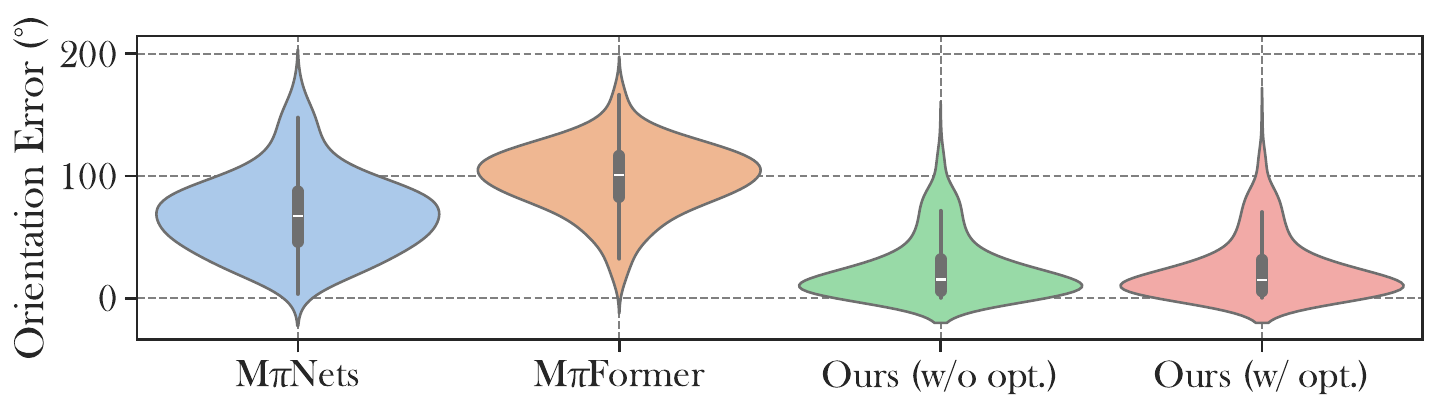} 
        \hfill
        \text{(d) orientation target error between end effector pose and given pose} \\[0.5em]
        \label{fig:goal_reaching_ori_error}
    \end{minipage}
    
    \caption{Analysis of errors in trajectories generated by different methods. (a) the minimum distance between the final end effector of generated motion and the manipulated object across all grasping tasks. (b) the horizontal distance between the centers of the bounding boxes of the object at final position and the target area across all placement tasks. (c) the position target error between the pose of final end effector and the given pose across all goal-reaching tasks. (d) the orientation target error between the pose of final end effector and the given pose across all goal-reaching tasks.}
    \label{fig:failure_case_analysis}
\end{figure}

\subsection{Results Analysis} 
As illustrated in \cref{fig:success_motion_generation}, we visualize the trajectories generated by our method to successfully grasp and place objects. We conduct a comprehensive evaluation of the performance of three models on various testing sets. As shown in \cref{table:quantitative_comparison}, \model demonstrates the highest success rates across the three mobile manipulation tasks. Specifically, for seen environments during training phase, \model achieves the best success rates of 30.54\% in object grasping, 22.89\% in object placement, and 30.49\% in goal-reaching. However, the success rates of \mpinets and \mpiformer did not exceed 5\% in any task. For tasks in unknown 3D scenes, \model also demonstrates stronger generalizability for novel environments than two baselines, with success rates of 14.33\%, 12.25\%, and 12.61\% across three mobile manipulation tasks.

Additionally, we find that the trajectories generated by the baselines generally bring the end effector close to the manipulated object (see \cref{fig:failure_case_analysis}\redtext{a}), target placement area (see \cref{fig:failure_case_analysis}\redtext{b}) or target end effector goal (see \cref{fig:failure_case_analysis}\redtext{c} and see \cref{fig:failure_case_analysis}\redtext{d}). However, the inherent shortsightedness of step-by-step autoregressive planning commonly prevents the end effector from successfully converging to an effective grasping pose (see \cref{fig:failed_motion_generation}\redtext{a}), precise placement (see \cref{fig:failed_motion_generation}\redtext{b}) or reaching the target goal (see \cref{fig:failed_motion_generation}\redtext{c}). In contrast, \model directly generates and optimizes an entire whole-body motion through an iterative denoising process, thus avoiding these issues, though this comes with an increased solving time.

Furthermore, we observe that frequent jittery movements of the end effector during the transition between adjacent states are the most common failure factor for the baselines. This issue arises mainly due to the frequent distributional shifts and data variance in model training, as well as the inherent shortsightedness of the autoregressive process, which typically results in an ambiguous decision boundary. Autoregressive planning considers only the partially observed sequence of states, without accounting for the global goal of the task. This shortsightedness results in suboptimal planning, especially in long-horizon and high-dimensional mobile manipulation, where the final end effector pose often deviates from the optimal solution, compromising the overall task performance.

The \model has three advantages over the previous SOTA neural motion planners in handling complex mobile manipulation tasks with the 3D environments.

\begin{table}[t!]
    \centering
    \tiny
    \caption{Physical metrics of trajectories generated by different methods across three mobile manipulation tasks.}
    \setlength{\tabcolsep}{3pt}
    \setstretch{1}
    \begin{tabular}{cc cccccc}
        \toprule
        \makecell{\makebox[0.5cm][c]{Test Set}} & \makecell{\makebox[1.7cm][c]{Methods}} & \makecell[c]{Coll. \\ Rate (\%)} & \makecell[c]{Avg. Coll. \\ Depth (cm)} &\makecell[c]{Med. Coll. \\ Depth (cm)} & \makecell[c]{Avg. Config \\ \ac{sparc}} & \makecell[c]{Avg. End Eff \\ \ac{sparc}} & \makecell[c]{Joint \\ Viol. (\%)} \\
        \midrule
        \rowcolor{MistyRose}
        \cellcolor{white} & \mpinets & $26.40$ & $5.03$ & $3.65$ & $-4.55$ & $-2.72$ & $5.59$ \\
        \rowcolor{MistyRose}
        \cellcolor{white} & \mpiformer & $37.06$ & $4.71$ & $2.94$ & $-4.35$ & $-2.42$ & $1.86$ \\
        \rowcolor{MistyRose}
        \cellcolor{white} & Ours (w/o opt.) & $34.27$ & $3.67$ & $2.34$ & $-2.69$ & $-2.22$ & $0.21$ \\
        \rowcolor{MistyRose}
        \cellcolor{white} & Ours (w/ opt.) & $20.29$ & $3.77$ & $2.50$ & $-3.29$ & $-2.94$ & $0.41$ \\

        \cmidrule{2-8}

        \rowcolor{LightCyan1}
        \cellcolor{white} & \mpinets & $32.44$ & $5.81$ & $4.04$ & $-6.87$ & $-3.16$ & $31.00$ \\
        \rowcolor{LightCyan1}
        \cellcolor{white} & \mpiformer & $63.33$ & $5.46$ & $3.23$ & $-6.81$ & $-3.20$ & $30.00$ \\
        \rowcolor{LightCyan1}
        \cellcolor{white} & Ours (w/o opt.) & $42.44$ & $4.58$ & $2.56$ & $-2.69$ & $-2.40$ & $0.78$ \\
        \rowcolor{LightCyan1}
        \cellcolor{white} & Ours (w/ opt.) & $24.78$ & $5.60$ & $3.68$ & $-3.90$ & $-2.63$ & $0.33$ \\

        \cmidrule{2-8}
        
        \rowcolor{LightOliveGreen}
        \cellcolor{white} & \mpinets & $21.22$ & $5.61$ & $4.33$ & $-4.56$ & $-2.55$ & $0.69$ \\
        \rowcolor{LightOliveGreen}
        \cellcolor{white} & \mpiformer & $71.86$ & $6.78$ & $4.74$ & $-4.67$ & $-3.57$ & $0.00$ \\
        \rowcolor{LightOliveGreen}
        \cellcolor{white} & Ours (w/o opt.) & $27.06$ & $4.09$ & $2.72$ & $-2.70$ & $-2.29$ & $0.20$ \\
        \rowcolor{LightOliveGreen}
        \multirow{-12}{*}{
            \makebox[0.5cm][c]{\rotatebox[origin=c]{90}{
                \cellcolor{white}
                \makebox[0.5cm][c]{Test Set One (seen scenes)}
        }}} & Ours (w/ opt.) & $20.20$ & $4.24$ & $3.04$ & $-2.80$ & $-2.39$ & $0.29$ \\

        \midrule
        \midrule


        \rowcolor{MistyRose}
        \cellcolor{white} & \mpinets & $33.73$ & $3.22$ & $1.47$ & $-4.96$ & $-2.87$ & $3.88$ \\
        \rowcolor{MistyRose}
        \cellcolor{white} & \mpiformer & $32.54$ & $1.96$ & $1.40$ & $-4.67$ & $-2.54$ & $0.60$ \\
        \rowcolor{MistyRose}
        \cellcolor{white} & Ours (w/o opt.) & $53.13$ & $4.79$ & $2.42$ & $-2.61$ & $-2.01$ & $0.30$ \\
        \rowcolor{MistyRose}
        \cellcolor{white} & Ours (w/ opt.) & $36.42$ & $5.47$ & $2.62$ & $-3.16$ & $-2.62$ & $0.30$ \\

        \cmidrule{2-8}

        \rowcolor{LightCyan1}
        \cellcolor{white} & \mpinets & $37.89$ & $6.60$ & $4.37$ & $-6.30$ & $-3.36$ & $33.33$ \\
        \rowcolor{LightCyan1}
        \cellcolor{white} & \mpiformer & $66.10$ & $6.08$ & $3.63$ & $-6.65$ & $-3.32$ & $15.67$ \\
        \rowcolor{LightCyan1}
        \cellcolor{white} & Ours (w/o opt.) & $57.83$ & $8.75$ & $6.80$ & $-2.51$ & $-2.24$ & $0.85$ \\
        \rowcolor{LightCyan1}
        \cellcolor{white} & Ours (w/ opt.) & $27.68$ & $9.22$ & $7.70$ & $-4.06$ & $-3.76$ & $0.28$ \\

        \cmidrule{2-8}
        
        \rowcolor{LightOliveGreen}
        \cellcolor{white} & \mpinets & $20.90$ & $7.64$ & $6.46$ & $-4.62$ & $-2.75$ & $0.36$ \\
        \rowcolor{LightOliveGreen}
        \cellcolor{white} & \mpiformer & $49.19$ & $7.60$ & $5.45$ & $-4.62$ & $-3.56$ & $0.00$ \\
        \rowcolor{LightOliveGreen}
        \cellcolor{white} & Ours (w/o opt.) & $50.45$ & $6.09$ & $3.65$ & $-2.65$ & $-2.17$ & $0.00$ \\
        \rowcolor{LightOliveGreen}
        \multirow{-12}{*}{
            \makebox[0.5cm][c]{\rotatebox[origin=c]{90}{
                \cellcolor{white}
                \makebox[0.5cm][c]{Test Set Two (unseen scenes)}
        }}} & Ours (w/ opt.) & $43.60$ & $6.53$ & $4.29$ & $-2.70$ & $-2.26$ & $0.00$ \\

        \bottomrule
    \end{tabular}%
    \begin{tablenotes}
        \footnotesize
        \item We calculate a series of physical metrics for the trajectories generated by different models. These metrics include the overall collision rate, the average and median collision depths with the environment for all collision-planning cases, the average \ac{sparc} values in both the configuration space and end effector space, as well as overall joint violation rate across all mobile manipulation tasks.
    \end{tablenotes}
    \label{table:physical_comparison}
\end{table}

\begin{figure*}[t!]
    \centering
    \includegraphics[width=\linewidth]{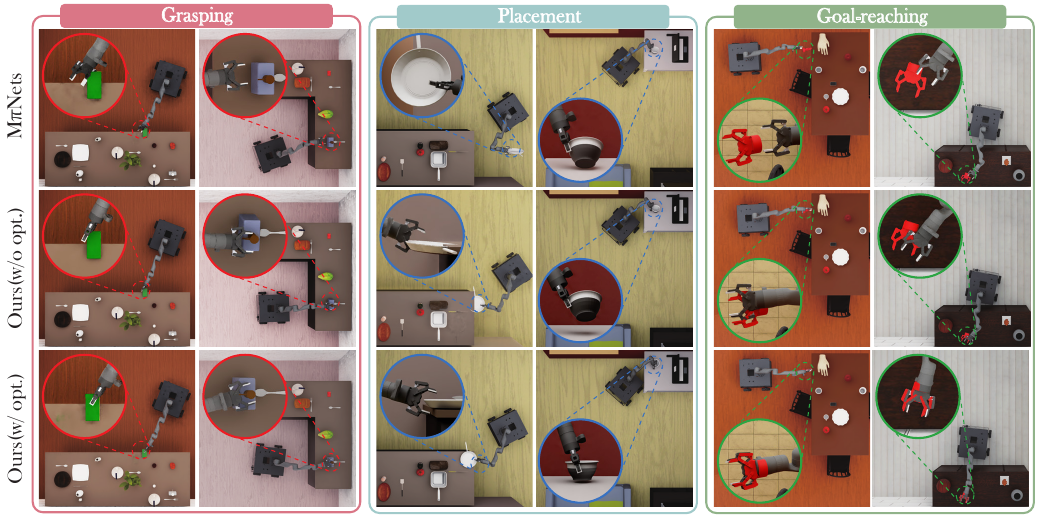}
    \caption{Final states of generated trajectory: \model (Ours) \vs \mpinets (Baseline). This illustration, rendered in NVIDIA Isaac Sim, displays the final states of trajectories generated by various methods across three mobile manipulation tasks, highlighting the comparative performance of \model and \mpinets in task completion.}
    \label{fig:quality_comparison}
\end{figure*}

\begin{table*}[t!]
    \centering
    \small
    \caption{Success rate (\%) for grasping and placing various objects using different methods.}
    \resizebox{\textwidth}{!}{%
        \setlength{\tabcolsep}{3pt}
        \setstretch{1.1}
        \begin{tabular}{cc cccccccccccccccc}
            \toprule
            \multirow{1}[1]{*}{Test Set} &  \multirow{1}[1]{*}{Methods} & \multirow{1}[1]{*}{All} &  
            \includegraphics[width=0.025\textwidth]{figures/object-images/Pan.pdf} & 
            \includegraphics[width=0.025\textwidth]{figures/object-images/Book.pdf} & \includegraphics[width=0.025\textwidth]{figures/object-images/Fork.pdf} & \includegraphics[width=0.025\textwidth]{figures/object-images/Knife.pdf} & \includegraphics[width=0.025\textwidth]{figures/object-images/Spoon.pdf} & \includegraphics[width=0.025\textwidth]{figures/object-images/Cup.pdf} & \includegraphics[width=0.025\textwidth]{figures/object-images/Bowl.pdf} & \includegraphics[width=0.025\textwidth]{figures/object-images/DishSponge.pdf} & \includegraphics[width=0.030\textwidth]{figures/object-images/Bottle.pdf} & \includegraphics[width=0.025\textwidth]{figures/object-images/Shaker.pdf} & \includegraphics[width=0.025\textwidth]{figures/object-images/Spatula.pdf} & \includegraphics[width=0.025\textwidth]{figures/object-images/Mug.pdf} & \includegraphics[width=0.020\textwidth]{figures/object-images/Egg.pdf} & \includegraphics[width=0.030\textwidth]{figures/object-images/Statue.pdf} & \includegraphics[width=0.030\textwidth]{figures/object-images/Plate.pdf}\\
            & & & Pan & Book & Fork & Knife & Spoon & Cup & Bowl & Sponge & Bottle & Shaker & Spatula & Mug & Egg & Statue & Plate \\
            \midrule
            \rowcolor{MistyRose}
            \cellcolor{white} & \mpinets & $0.00$ & $0.00$ & $0.00$ & $0.00$ & $0.00$ & $0.00$ & $0.00$ & $0.00$ & $0.00$ & $0.00$ & $0.00$ & $0.00$ & $0.00$ & $0.00$ & $0.00$ & $0.00$  \\
            \rowcolor{MistyRose}
            \cellcolor{white} & \mpiformer & $3.93$ & $3.57$ & $18.31$ & $0.00$ & $8.82$ & $2.50$ & $3.03$ & $4.76$ & $1.28$ & $3.10$ & $1.30$ & $0.00$ & $0.00$ & $0.00$ & $0.00$ & $6.98$ \\
            \rowcolor{MistyRose}
            \cellcolor{white} & Ours (w/o opt.) & $21.95$ & $32.14$ & $\underline{\bm{30.99}}$ & $0.00$ & $8.82$ & $13.75$ & $15.15$ & $14.29$ & $21.79$ & $\underline{\bm{24.81}}$ & $28.57$ & $4.65$ & $18.37$ & $\underline{\bm{12.50}}$ & $26.58$ & $27.13$ \\
            \rowcolor{MistyRose}
            \cellcolor{white} & Ours (w/ opt.) & $\underline{\bm{30.54}}$ & $\underline{\bm{39.29}}$ & ${28.17}$ & $\underline{\bm{6.45}}$ & $\underline{\bm{11.76}}$ & $\underline{\bm{{20.00}}}$ & $\underline{\bm{27.27}}$ & $\underline{\bm{23.81}}$ & $\underline{\bm{30.77}}$ & ${22.48}$ & $\underline{\bm{{37.84}}}$ & $\underline{\bm{18.60}}$ & $\underline{\bm{28.57}}$ & $\underline{\bm{12.50}}$ & $\underline{\bm{{55.70}}}$ & $\underline{\bm{45.74}}$ \\
            
            \cmidrule{2-18}
            
            \rowcolor{LightCyan1}
            \cellcolor{white} & \mpinets & $2.33$ & $0.00$ & $3.90$ & $0.00$ & $1.69$ & $0.00$ & $0.00$ & $4.00$ & $1.80$ & $4.76$ & $0.00$ & $0.00$ & $\backslash$ & $0.00$ & $0.00$ & $6.67$ \\
            \rowcolor{LightCyan1}
            \cellcolor{white} & \mpiformer & $0.89$ & $6.82$ & $1.30$ & $0.00$ & $0.00$ & $0.00$ & $0.00$ & $0.00$ & $0.00$ & $0.95$ & $0.00$ & $0.00$ & $\backslash$ & $0.00$ & $0.00$ & $3.33$ \\
            \rowcolor{LightCyan1}
            \cellcolor{white} & Ours (w/o opt.) & $4.67$ & $9.09$ & $16.88$ & $3.85$ & $1.69$ & $0.00$ & $0.00$ & $8.00$ & $0.00$ & $2.86$ & $0.00$ & $3.47$ & $\backslash$ & $0.00$ & $8.00$ & $8.89$ \\ 
            \rowcolor{LightCyan1}
            \multirow{-8}{*}{
                \makebox[0.8cm][c]{\rotatebox[origin=c]{90}{
                    \cellcolor{white}
                    \makebox[2cm][c]{Test Set One}
            }}} & Ours (w/ opt.) & $\underline{\bm{22.89}}$ & $\underline{\bm{36.36}}$ & $\underline{\bm{64.94}}$ & $\underline{\bm{3.85}}$ & $\underline{\bm{3.39}}$ & $\underline{\bm{4.41}}$ & $\underline{\bm{15.38}}$ & $\underline{\bm{32.00}}$ & $\underline{\bm{28.83}}$ & $\underline{\bm{3.81}}$ & $\underline{\bm{0.00}}$ & $\underline{\bm{11.81}}$ & $\backslash$ & $\underline{\bm{33.33}}$ & $\underline{\bm{20.00}}$ & $\underline{\bm{58.89}}$ \\

            \midrule
            \midrule
            
            \rowcolor{MistyRose}
            \cellcolor{white} & \mpinets & $0.00$ & $0.00$ & $0.00$ & $0.00$ & $0.00$ & $0.00$ & $0.00$ & $0.00$ & $0.00$ & $0.00$ & $0.00$ & $0.00$ & $0.00$ & $0.00$ & $\backslash$ & $\backslash$ \\
            \rowcolor{MistyRose}
            \cellcolor{white} & \mpiformer & $3.28$ & $4.76$ & $0.00$ & $0.00$ & $0.00$ & $0.00$ & $\underline{\bm{11.76}}$ & $\underline{\bm{19.23}}$ & $4.76$ & $0.00$ & $0.00$ & $0.00$ & $0.00$ & $0.00$ & $\backslash$ & $\backslash$ \\
            \rowcolor{MistyRose}
            \cellcolor{white} & Ours (w/o opt.) & $9.25$ & $9.52$ & $11.76$ & $0.00$ & $0.00$ & $\underline{\bm{24.00}}$ & $0.00$ & $7.69$ & $19.05$ & $\underline{\bm{3.85}}$ & $23.81$ & $9.52$ & $\underline{\bm{23.81}}$ & $\underline{\bm{4.76}}$ & $\backslash$ & $\backslash$ \\
            \rowcolor{MistyRose}
            \cellcolor{white} & Ours (w/ opt.) & $\underline{\bm{14.33}}$ & $\underline{\bm{38.10}}$ & $\underline{\bm{23.53}}$ & $\underline{\bm{7.69}}$ & $\underline{\bm{4.76}}$ & ${4.00}$ & $8.82$ & $3.85$ & $\underline{\bm{33.33}}$ & ${0.00}$ & $\underline{\bm{38.10}}$ & $\underline{\bm{33.33}}$ & $19.05$ & $\underline{\bm{4.76}}$ & $\backslash$ & $\backslash$ \\
            
            \cmidrule{2-18}
            
            \rowcolor{LightCyan1}
            \cellcolor{white} & \mpinets & $0.85$ & $\backslash$ & $6.25$ & $0.00$ & $\backslash$ & $0.00$ & $0.00$ & $0.00$ & $\backslash$ & $0.00$ & $0.00$ & $0.00$ & $0.00$ & $0.00$ & $\backslash$ & $\backslash$ \\
            \rowcolor{LightCyan1}
            \cellcolor{white} & \mpiformer & $0.28$ & $\backslash$ & $2.08$ & $0.00$ & $\backslash$ & $0.00$ & $0.00$ & $0.00$ & $\backslash$ & $0.00$ & $0.00$ & $0.00$ & $0.00$ & $0.00$ & $\backslash$ & $\backslash$ \\
            \rowcolor{LightCyan1}
            \cellcolor{white} & Ours (w/o opt.) & $5.70$ & $\backslash$ & $25.00$ & $0.00$ & $\backslash$ & $2.08$ & $0.00$ & $16.67$ & $\backslash$ & $0.00$ & $0.00$ & $\underline{\bm{12.50}}$ & $0.00$ & $0.00$ & $\backslash$ & $\backslash$ \\
            \rowcolor{LightCyan1}
            \multirow{-8}{*}{
                \makebox[0.8cm][c]{\rotatebox[origin=c]{90}{
                    \cellcolor{white}
                    \makebox[2cm][c]{Test Set Two}
            }}} & Ours (w/ opt.) & $\underline{\bm{12.25}}$ & $\backslash$ & $\underline{\bm{62.50}}$ & $0.00$ & $\backslash$ & $0.00$ & $\underline{\bm{14.58}}$ & $\underline{\bm{50.00}}$ & $\backslash$ & $0.00$ & $0.00$ & $4.17$ & $0.00$ & $0.00$ & $\backslash$ & $\backslash$ \\
 
            \bottomrule
        \end{tabular}%
    }%
    \begin{tablenotes}
        \footnotesize
        \item We report the success rates of \model and two baselines for \textcolor{grasping}{grasping} and \textcolor{placement}{placing} 15 objects from the test set. Test set one includes the results from testing in familiar scenes that were encountered during training, while test set two reflects the results in novel scenes that were not seen during training.
    \end{tablenotes}
    \label{table:success_rate_of_diff_objects}
\end{table*}

\textbf{a) Motion generation with trajectory optimization:} Effective mobile manipulation necessitates robots to interact with their surroundings, while adhering to multiple constraints imposed by both the agent embodiment and the environmental context. Unlike baselines, \model avoids the paradigm to design explicit loss functions during model training for learning certain constraints (\eg, collision avoidance~\cite{fishman2022mpinets}). Instead, \model integrates physical constraints into the diffusion model and design a guided optimization mechanism in the generation process, leading to an efficient way to introduce implicit task requirements in a differentiable manner for jointly optimizing the task goal sampling and the trajectory generation, ensuring physical plausibility and task-related completeness of the generated motion. This design also facilitates the integration of multiple constraints and fine-tuning of hyperparameters.

As shown in \cref{table:physical_comparison}, \model without tajectory optimization exhibits a higher collision rate with the seen environment compared to \mpinets, because collision-avoidance constraint was not considered during the its training phase. However, by introducing the collision-avoidance guided function defined in \cref{eq:collision_cost}, the optimized trajectories demonstrate the lowest collision rates on the familiar 3D scenes. Similarly, as indicated in 6th and 7th columns of \cref{table:physical_comparison}, the \ac{sparc} value of the trajectories sampled by \model decreases with the guidance of \cref{eq:traj_smooth_cost}. These smooth trajectories benefits the action execution for the low-level controller. Moreover, our optimization framework also supports data-driven objective functions. As illustrated in \cref{fig:quality_comparison}, by incorporating implicit grasping and placement energy functions, the task completion of the generated trajectories for grasping and placing objects significantly improves. 

Abovementioned optimization framework not only enhances the adaptability of learned planner to complex tasks but also outperforms baseline methods in ensuring safe and successful manipulations in various scenarios.

\textbf{b) Generalizability and robustness in diverse environments:} 
\model utilizes a robot-centric 3D scan for visual observation, which enhances its generalizability across diverse scenarios compared to scene-centric models. By focusing on the local environment around the robot rather than the entire scene, \model is more readily extensible to unknown and real-world scenarios.
It has demonstrated robust performance in a variety of settings. As evidenced by the results in \cref{table:quantitative_comparison}, \model achieves success rates of 14.33\%, 12.25\%, and 12.61\% for three evaluation tasks, respectively, in previously unseen scenes. Furthermore, when deployed in real household environments (see \cref{sec:real_exp}), the model trained on simulated data can be directly applied to real 3D environments without any performance gap. Furthermore, our method exhibits strong generalizability at the object level, facilitating the manipulation of a diverse range of geometric shapes and object categories. As demonstrated in \cref{table:success_rate_of_diff_objects}, our approach significantly surpasses baseline models in terms of success rates for grasping and placing various objects.

\begin{figure}[t!]
    \footnotesize
    \centering
    \begin{minipage}{\linewidth}
        \centering
        \includegraphics[width=\textwidth]{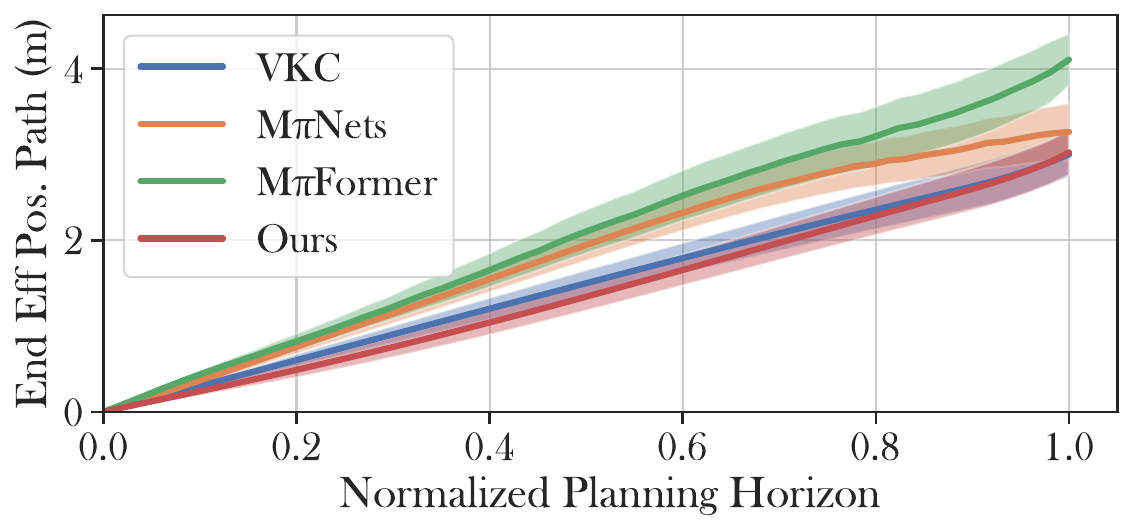} 
        \hfill
        \text{(a) end effector position paths generated via different planners} \\[0.5em]
        \label{fig:pos_path}
    \end{minipage}

    \begin{minipage}{\linewidth}
        \centering
        \includegraphics[width=\textwidth]{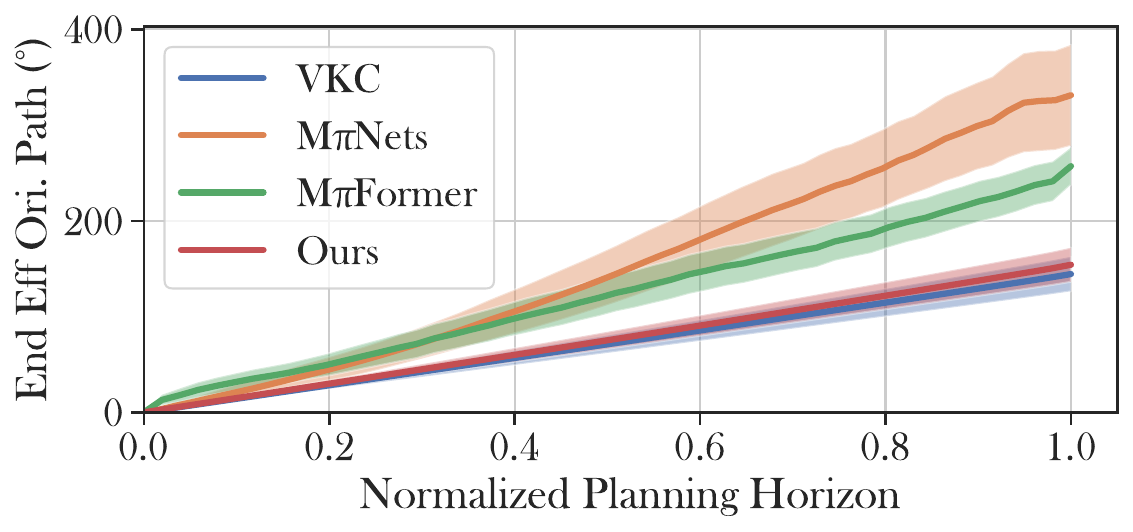} 
        \hfill
        \text{(b) end effector orientation paths generated via different planners} \\[0.5em]
        \label{fig:ori_path}
    \end{minipage}
    
    \caption{Performance comparison of different models in learning globally optimal expert planner. We summarize the planning results of goal-reaching tasks, where \mpinets, \mpiformer and the expert planner (VKC) all successfully perform. Then, we plot the continuous path curves for position and orientation of the end effector as the planning progresses. In the path curves, each point represents the average path length of position or orientation traversed by the end effector at the current normalized planning step, while the shaded area indicates the variance of the path length. Obviously, the planning results of \model closely align with those of the globally optimal expert planner.}
    \label{fig:com_for_learn_mm}
\end{figure}

\textbf{c) Near optimal trajectory generation:} 
\model generates trajectories via an iterative denoising process, which infers the entire action sequence  rather than only single-step actions. This property inherently promotes global optimality, as it considers the long-term effects of each action throughout the sequence, avoiding the pitfalls of myopic planning. This is crucial for high-dimensional mobile manipulation, where even small errors can be costly (see the 1st row of \cref{fig:quality_comparison}). Consequently, the trajectories produced by \model not only exhibit temporal consistency but also closely align with the globally optimal paths planned by the expert planner, as shown in \cref{fig:com_for_learn_mm}. These results highlight the superiority of \model in learning high-dimensional, whole-body mobile manipulation with a focus on global optimization.

\subsection{Ablation Experiments}
To confirm the significance of the learned diffusion priors for trajectory sampling and optimization, we present both physical and task-related performance comparison for trajectories sampled with and without diffusion priors. As demonstrated in \cref{table:ablation_exp}, our method considerably outperforms the direct use of inverse Langevin diffusion in terms of convergence speed and success rate. By learning a prior trajectory-level generator, \model efficiently guides the optimization algorithm to rapidly search robust and high-quality solutions within a reduced search space. In our experiments, the sampling process of the inverse Langevin diffusion is defined by referencing to~\cite{urain2023se}, where
\begin{equation}
    \label{eq:langevin_diffusion}
    \begin{aligned}
    \boldsymbol{\tau}_{k-1}=\boldsymbol{\tau}_k + 0.5 \alpha_k^2 \nabla_{\boldsymbol{\tau}_k} \varphi(\boldsymbol{\tau}_k) + \alpha_k \boldsymbol{\epsilon}, \boldsymbol{\epsilon} \sim \mathcal{N}(\mathbf{0}, \mathbf{I}),
    \end{aligned}
\end{equation}
with pre-defined step dependent coefficient $\alpha_k$ and objective function $\varphi(\cdot)$ previously defined in \cref{eq:likelihood_task}.

\begin{table}[ht!]
    \centering
    \small
    \caption{Quantitative comparison of \model and inverse Langevin diffusion in solving goal-reaching task.}
    \setlength{\tabcolsep}{3pt}
    \begin{tabular}{cccc}
        \toprule
        Diffusion Priors & Iterative Step & Succ. Rate (\%) & Coll. Depth (m) \\
        \midrule
        \ding{55} & $50$ & $0.00$ & $7.15$ \\
        \ding{55} & $500$ & $0.00$ & $6.94$ \\
        \ding{55} & $1000$ & $0.00$ & $6.79$ \\
        \ding{55} & $2000$ & $0.00$ & $6.65$ \\
        \midrule
        \ding{51} & $\underline{\bm{50}}$ & $\underline{\bm{51.00}}$ & $\underline{\bm{4.79}}$ \\
        \bottomrule
    \end{tabular}
    \begin{tablenotes}
        \footnotesize
        \item We randomly select 100 goal-reaching tasks to test the performance of trajectory optimization with and without the learned diffusion priors.
    \end{tablenotes}
    \label{table:ablation_exp}
\end{table}

\begin{figure*}[t!]
    \centering
    \footnotesize
    \begin{minipage}{0.28\linewidth}
        \centering
        \includegraphics[width=\linewidth]{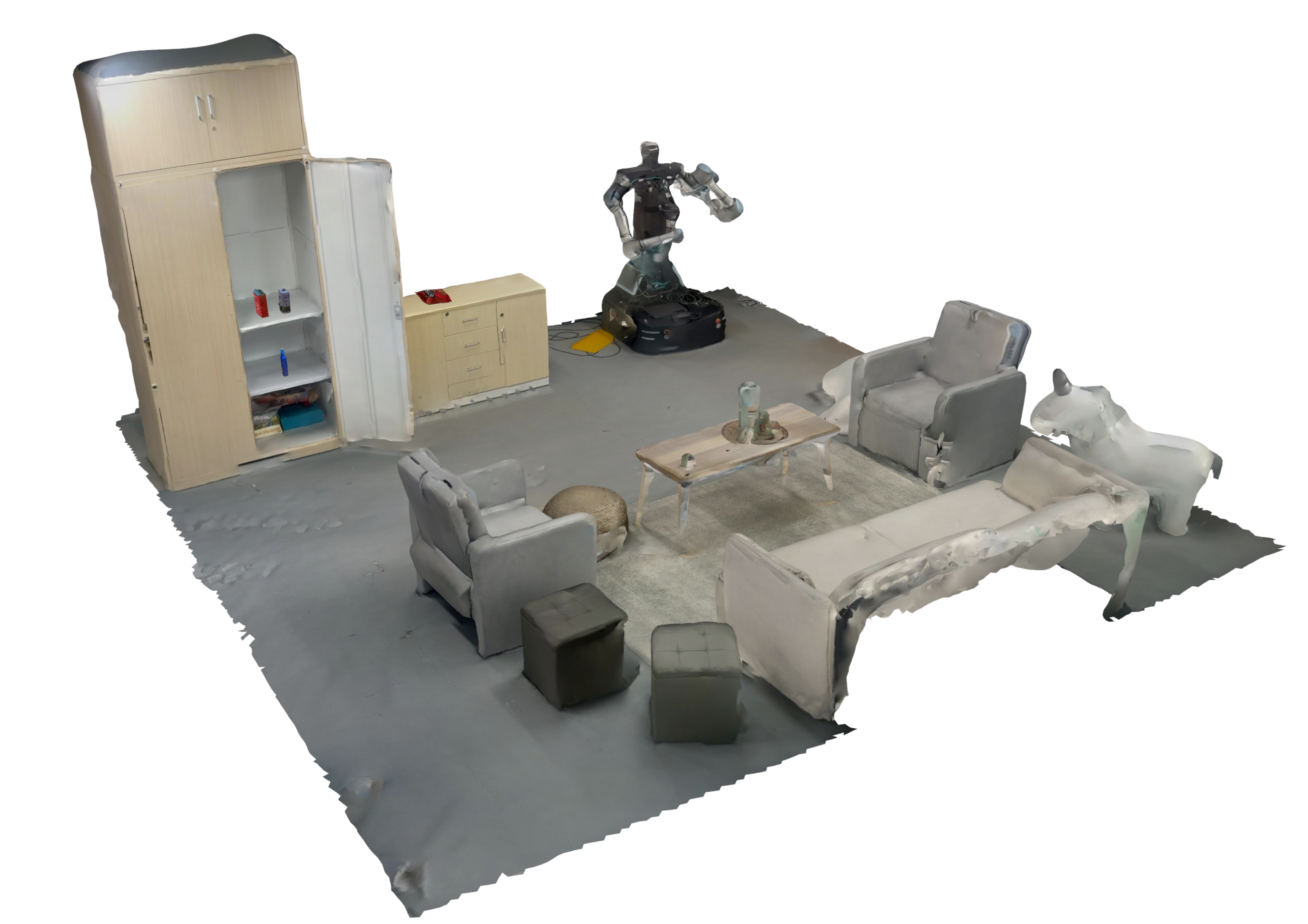} 
        \hfill
        \text{(a) real-world 3D scan}
        \label{fig:3D_scan}
    \end{minipage}
    \hfill
    \begin{minipage}{0.288\linewidth}
        \centering
        \includegraphics[width=\linewidth]{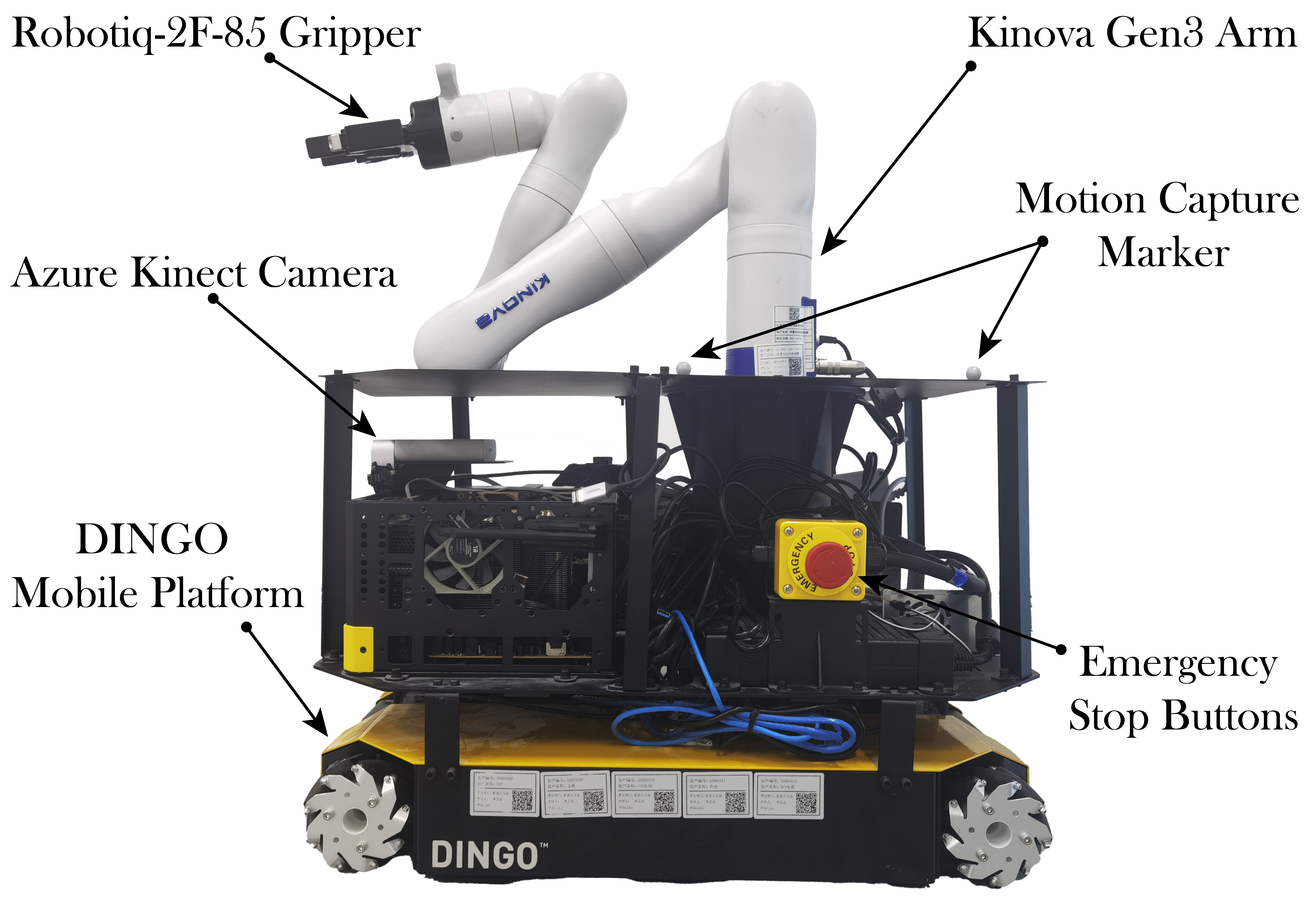} 
        \hfill
        \text{(b) mobile manipulator}
        \label{fig:real_robot}
    \end{minipage}
    \hfill
    \begin{minipage}{0.38\linewidth}
        \centering
        \includegraphics[width=\linewidth]{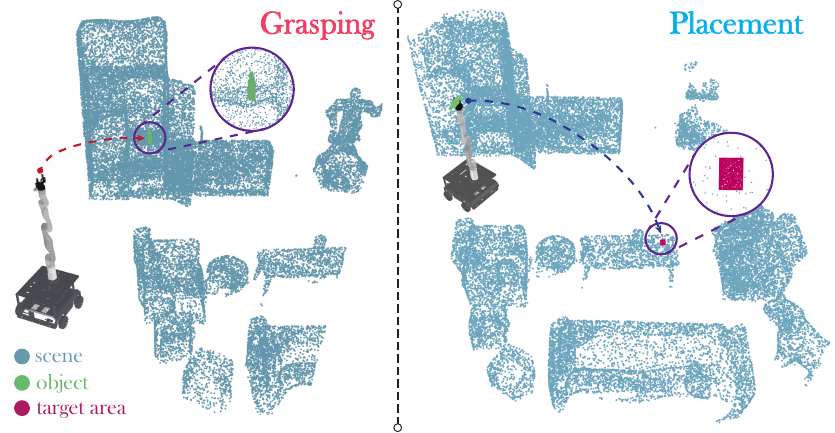} 
        \hfill
        \text{(c) robot-centic 3D scans for grasping and placement}
        \label{fig:real_robot_centric_3D_scan}
    \end{minipage}
    
    \caption{Real-world 3D environment and robot system. (a) The real-world 3D environment is scanned and reconstructed by using PolyCam software. (b) The mobile manipulation system comprises a 3-DoF mobile base, a 7-DoF manipulation arm, and additional attachments. (c) The robot-centric 3D scans are utilized for grasping and placements tasks in experiment one of \cref{fig:real_world_exp}.}
    \label{fig:real_world_env_robot}
\end{figure*}

\begin{figure*}[t!]
    \centering
    \includegraphics[width=\linewidth]{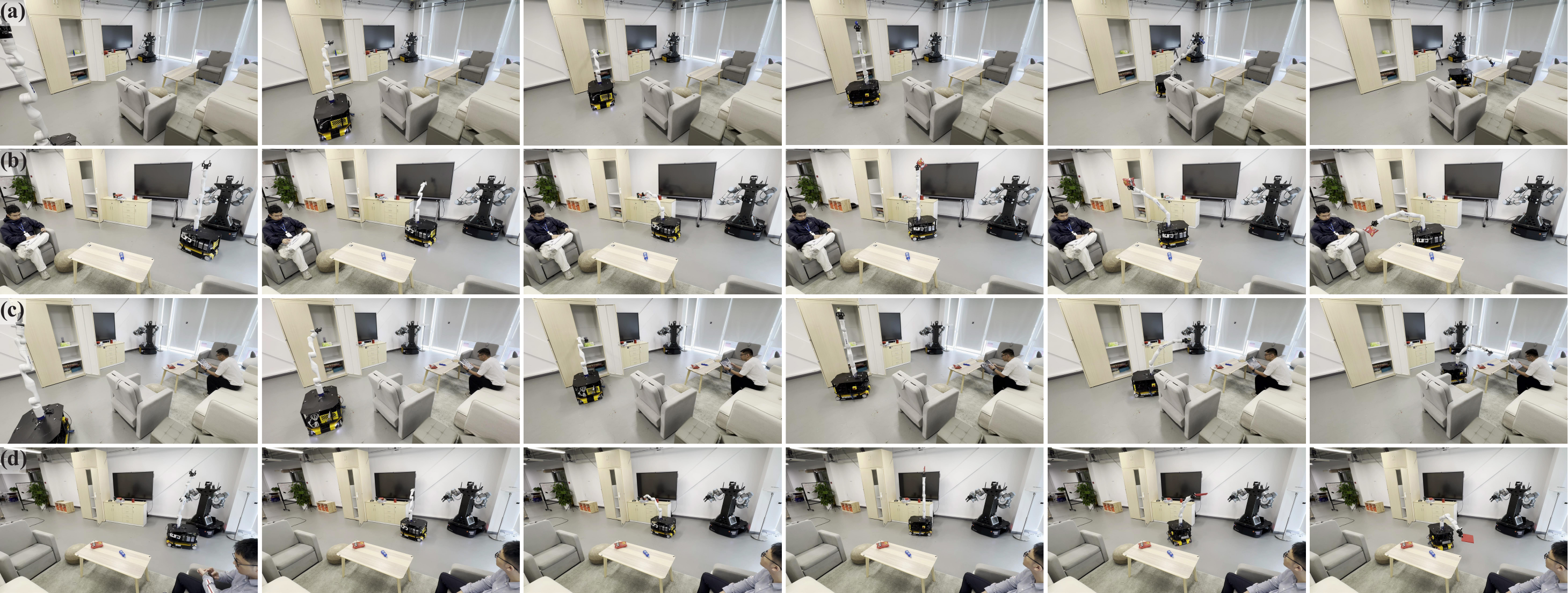}
    \caption{Real-world experiments. From top to bottom, the figures depict the mobile manipulator (a) taking a bottle from the cabinet and placing it on the table, (b) handing a chip bag to a seated person, (c) retrieving a tea box from the cabinet and setting it on the table, and (d) delivering a book to a seated person.}
    \label{fig:real_world_exp}
\end{figure*}

\section{Experiments in Real-world 3D Scenes} \label{sec:real_exp}
This section illustrates the application of our method to a real mobile manipulator performing objects rearrangement and handover tasks in a real household environment. The real-world 3D environment and our robot system are depicted in \cref{fig:real_world_env_robot}. To the best of our knowledge, this is the first study to directly apply an \ac{il}-based neural motion planner trained on simulated data to real-world mobile manipulation tasks. We confirm that our method can seamlessly transfer from simulation to the real world. The following subsections introduce robot system settings (in \cref{sec:robot_sys}), real-world experiment setup (in \cref{sec:real_exp_setup}), and real-world experiment results (in \cref{sec:real_world_eval}).

\subsection{Robot System Settings} \label{sec:robot_sys}
The robot used in our real-world experiments is a 10-DoF mobile manipulator (see \cref{fig:real_world_env_robot}\redtext{b}), which consists of a 3-DoF Dingo base with omnidirectional Mecanum wheels, a 7-DoF Kinova Gen3 arm, and a Robotiq-2F-85 gripper. It shares the same geometric embodiment as the robot model used in simulations and has the capacity to perform intricate mobile manipulation in man-made environments.

\subsection{Experiment Setup} \label{sec:real_exp_setup}
\paragraph{Mobile Manipulation Tasks} 
We set up a series of object rearrangement tasks involving three common geometric shapes: planar, cuboids, and cylindrical objects. In these tasks, the robot is required to pick objects from their initial locations and place them to specified target areas (\eg, on a table surface or around a person) based on the segmented scene's natural 3D scan and object segmented masks. To successfully complete the task, the robot must not only execute each grasping and placement accurately but also avoid collisions or other physical violations. A similar object rearrangement task was explored in~\cite{gu2022multi} and has been shown to be highly challenging in real-world environments. The experimental site is set up in a real living room environment with various objects and obstacles. The model only trained with simulated data will be directly used in the real-world setting without any fine-tuning.
\paragraph{Experiment Preparation} 
As shown in \cref{fig:real_world_env_robot}\redtext{a}, we scan and reconstruct the household environment. The reconstructed scene's point cloud is then segmented and cropped to serve as the visual input (see \cref{fig:real_world_env_robot}\redtext{c}) for \model. Additionally, we adopt the algorithm proposed by~\cite{wang2022dual} to calculate the \ac{sdf} of the reconstructed scene for collision-avoidance cost computation during trajectory optimization. In real-world experiments, the overhead VICON system provides real-time localization of the robot.

\subsection{Experiment Results} \label{sec:real_world_eval}
As shown in \cref{fig:real_world_exp}, we conduct a series of pick-and-place tasks involving various common objects, including a bottle, a chip bag, a book and a tea box. For each task, the \model first generates a trajectory for grasping the object and then plans a subsequent trajectory for placing it in a target area on the table or around a person. In real-world experiments, we attempt to directly apply the model trained on simulated data to pick-and-place tasks in real-world 3D scenes and unseen objects, achieving significant success. By leveraging robot-centric 3D scans as visual input, our model first achieves seamless sim-to-real transfer in learning-based mobile manipulation. However, previous works either failed to directly apply models trained in simulation to real-world scenarios or were limited to highly structured environments. Additionally, we also demonstrate the generalizability and robustness of the trajectory optimization framework in handling previously unseen environments and objects.

\section{Limitations and Future Work} \label{sec:limits_and_future_work}
The primary limitations of the \model include its slow training and inference speed and strong dependence on the objective function designs. More details are as follows:

\paragraph{Slow Training and Inference} 
The training and inference of \model are both slow due to the numerous iterative steps required for generating outputs, a common issue with diffusion models. As \model optimizes the sampled trajectories at each iterative denoising step, it is incompatible with sampling acceleration algorithms such as DDIM~\cite{song2021denoising}, which often compromise optimization performance. As shown in \cref{table:quantitative_comparison}, the introduction of optimized guidance terms results in a 5 to 10-fold decrease in inference speed for \model.

\paragraph{Strong Dependence on Objective Designs} 
\model optimizes the sampled trajectories, depending heavily on the design of the energy and cost functions, as well as meticulous hyper-parameter tuning. These functions can be either explicitly defined through heuristic designs or implicitly derived from data-driven models. However, for the multi-stage tasks in~\cite{chi2023diffusionpolicy} and the human-like skills in~\cite{chi2024universal}, task-related optimization often proves impractical due to the challenges on designing smooth objective functions.

In future work, we will attempt to solve the aforementioned limitations by exploring the latest advancements in diffusion model acceleration and loss guidance algorithms to reduce the number of inference steps required without sacrificing optimized performance, such as new noise schedules~\cite{chen2023importance} and LGD-MC~\cite{song2023lossguided}. 

\section{Conclusion} \label{sec:conclusion}
We proposed \model, the first scene-conditioned motion generator tailored for mobile manipulation in \ac{eai}. \model seamlessly integrates multiple physical constraints and task objective, and employs generative modeling techniques to directly generate highly coordinated whole-body motion trajectories with physical plausibility and task completion from natural 3D scans. We demonstrate that the \model outperforms previous SOTA neural motion planners by a large margin on various tasks, establishing its efficacy and flexibility. Furthermore, we also demonstrated that the diffusion-based planning paradigm, along with using robot-centric 3D scans as visual observation, can be more effective for the real-world generalization and deployment of mobile manipulation.

{\small
    \bibliographystyle{IEEEtran}
    \bibliography{reference}
}

\newpage 
\begin{IEEEbiography}[{\includegraphics[width=1in,height=1.25in,clip,keepaspectratio]{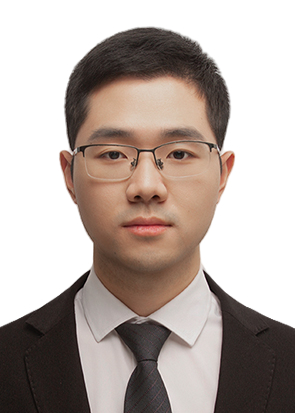}}]{Sixu Yan}
  received the M.S. degree from Shanghai Jiao Tong University (SJTU), Shanghai, China, in 2024, and the B.E. degree from Ocean University of China (OUC), Qingdao, China, in 2021, both in Mechanical Engineering. He is currently a first-year Ph.D. student at the School of Electronic Information and Communications, Huazhong University of Science and Technology (HUST). His research interests include robotics and computer vision.
\end{IEEEbiography}

\begin{IEEEbiography}[{\includegraphics[width=1in,height=1.25in,clip,keepaspectratio]{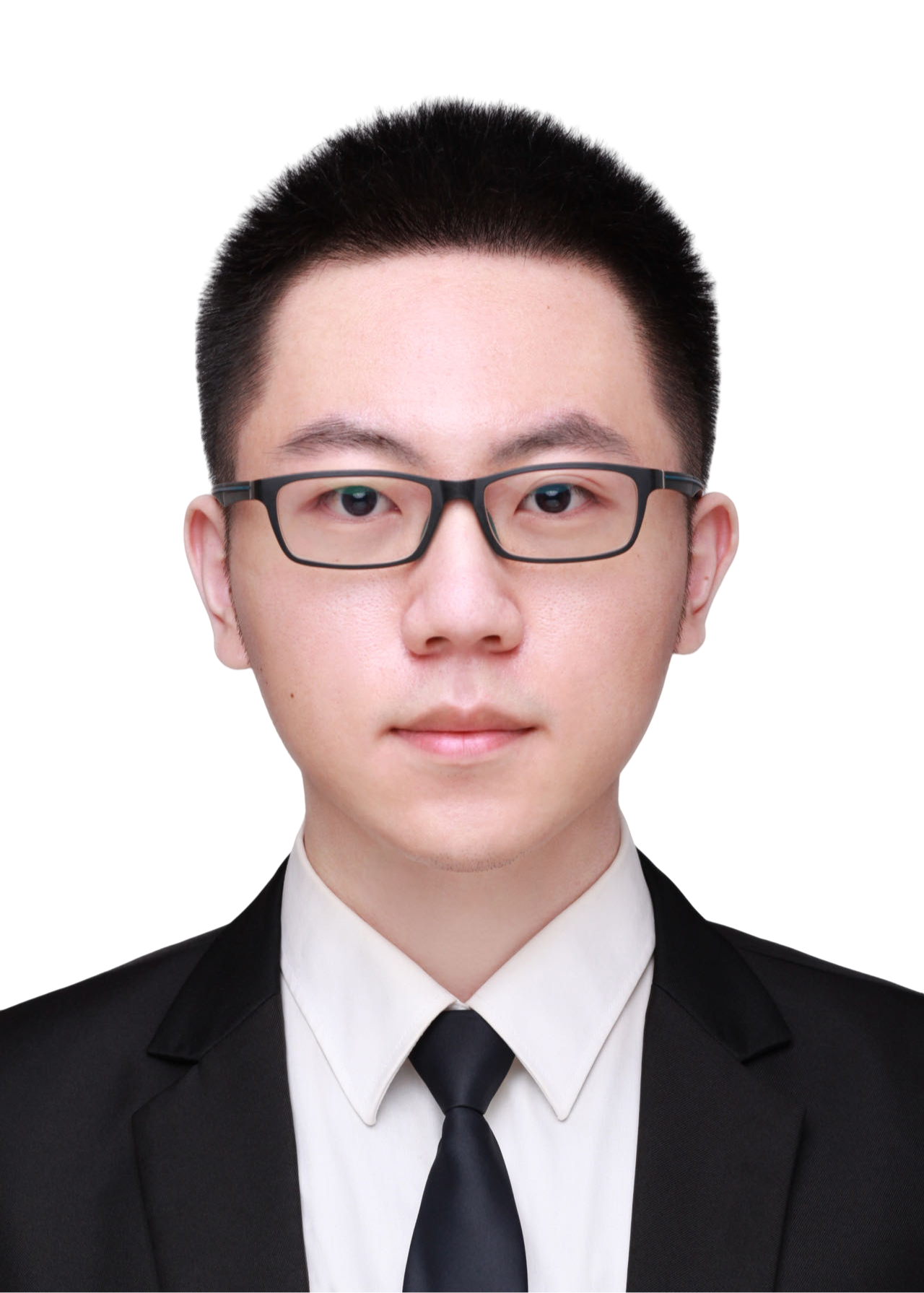}}]{Zeyu Zhang}
    (Member, IEEE) received his Ph.D. degree in Computer Science from the University of California, Los Angeles (UCLA) in 2023. He is currently a research scientist at State Key Laboratory of General Artificial Intelligence, Beijing Institute for General Artificial Intelligence (BIGAI). He received an M.S. degree in Computer Science from UCLA in 2019 and B.S. degree in Computer Science from Hunan University in 2017. His research interests focus on robot perception, learning, and cognitive robotics
\end{IEEEbiography}

\begin{IEEEbiography}[{\includegraphics[width=1in,height=1.25in,clip,keepaspectratio]{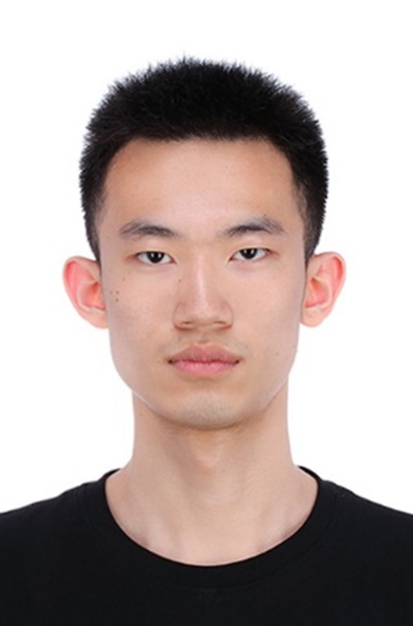}}]{Muzhi Han}
  received a B.E. in Mechanical Engineering from Tsinghua University, Beijing, China, in 2019. He is currently a final-year Ph.D. candidate at the University of California, Los Angeles (UCLA), advised by Prof. Song-Chun Zhu. His research interests include robotics and machine perception.
\end{IEEEbiography}

\begin{IEEEbiography}[{\includegraphics[width=1in,height=1.25in,clip,keepaspectratio]{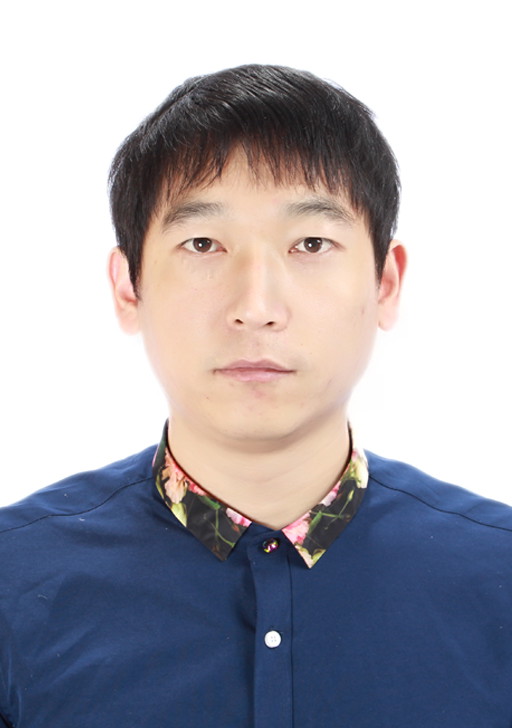}}]{Zaijin Wang}
    received the M.S. degree in Mechanical Engineering from North China University of Technology in 2013, and the B.E. degree in Mechanical Engineering from Qingdao University in 2010. He is currently a research engineer at the State Key Laboratory of General Artificial Intelligence, Beijing Institute for General Artificial Intelligence (BIGAI). His research interests include computer vision, machine learning, motion control, and cognitive robotics.
\end{IEEEbiography}

\begin{IEEEbiography}[{\includegraphics[width=1in,height=1.25in,clip,keepaspectratio]{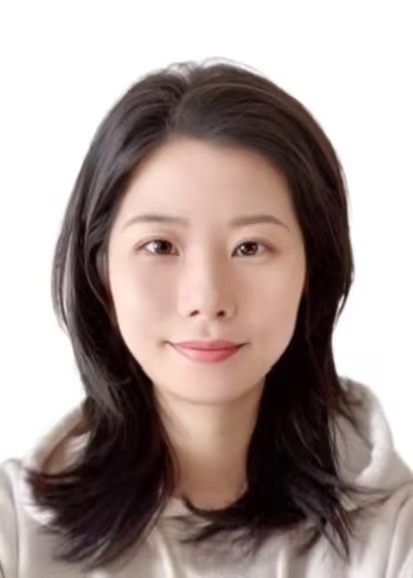}}]{Qi Xie}
    received the M.S. degree in Applied Data Science from University of Southern California in 2022. She is now a research engineer at State Key Laboratory of General Artificial Intelligence, Beijing Institute for General Artificial Intelligence (BIGAI). Her research interests include embodied AI and cognitive science.
\end{IEEEbiography}

\begin{IEEEbiography}[{\includegraphics[width=1in,height=1.25in,clip,keepaspectratio]{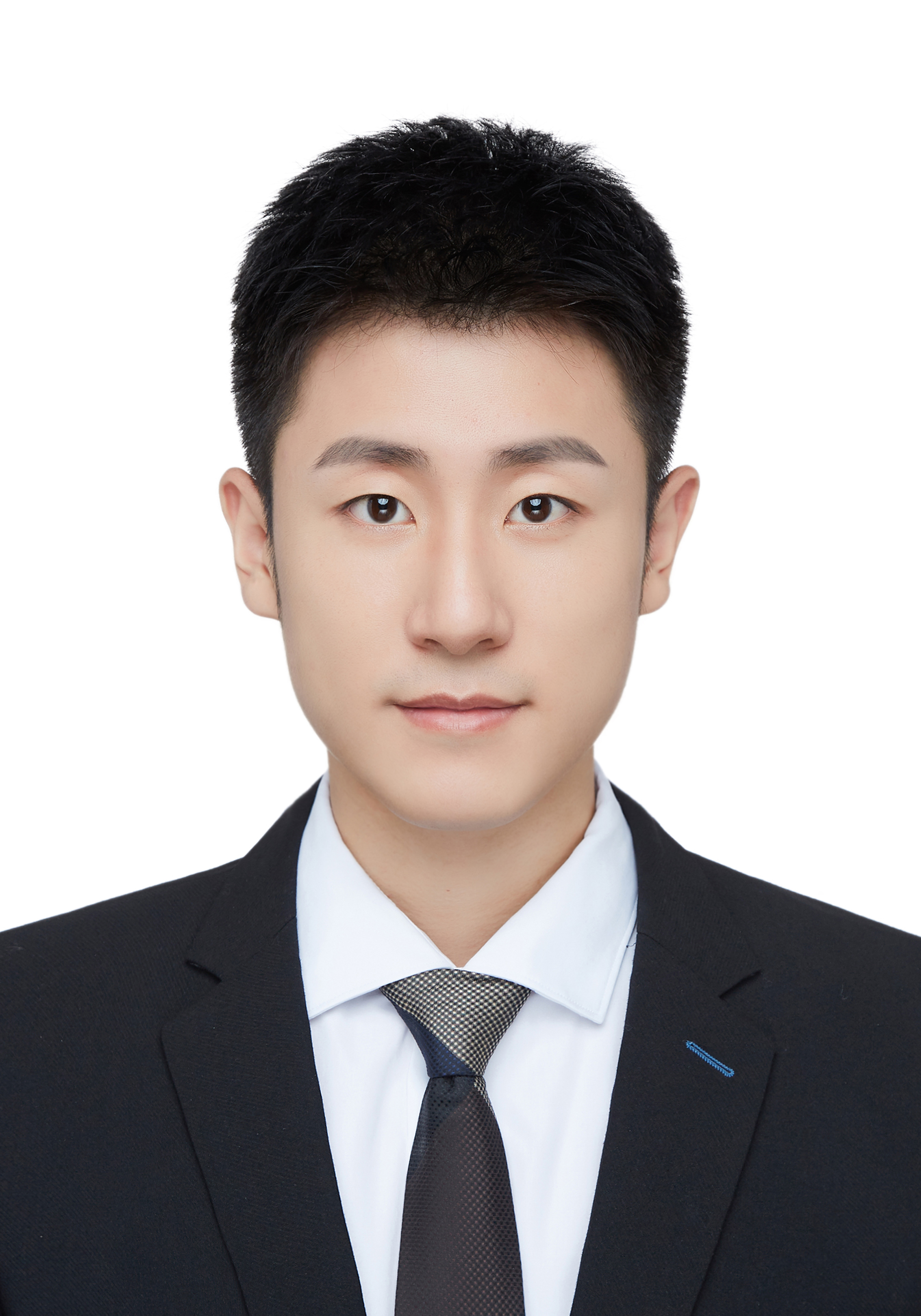}}]{Zhitian Li}
  received the M.S. degree in Mechanical Engineering from Beijing Institute of Technology (BIT), Beijing, China, in 2023, and the B.E. degree from BIT in 2020. He is currently a first-year Ph.D. student at the School of Automation Science and Electrical Engineering, Beihang University (BUAA). His research interests include robot planning and control.
\end{IEEEbiography}

\begin{IEEEbiography}[{\includegraphics[width=1in,height=1.25in,clip,keepaspectratio]{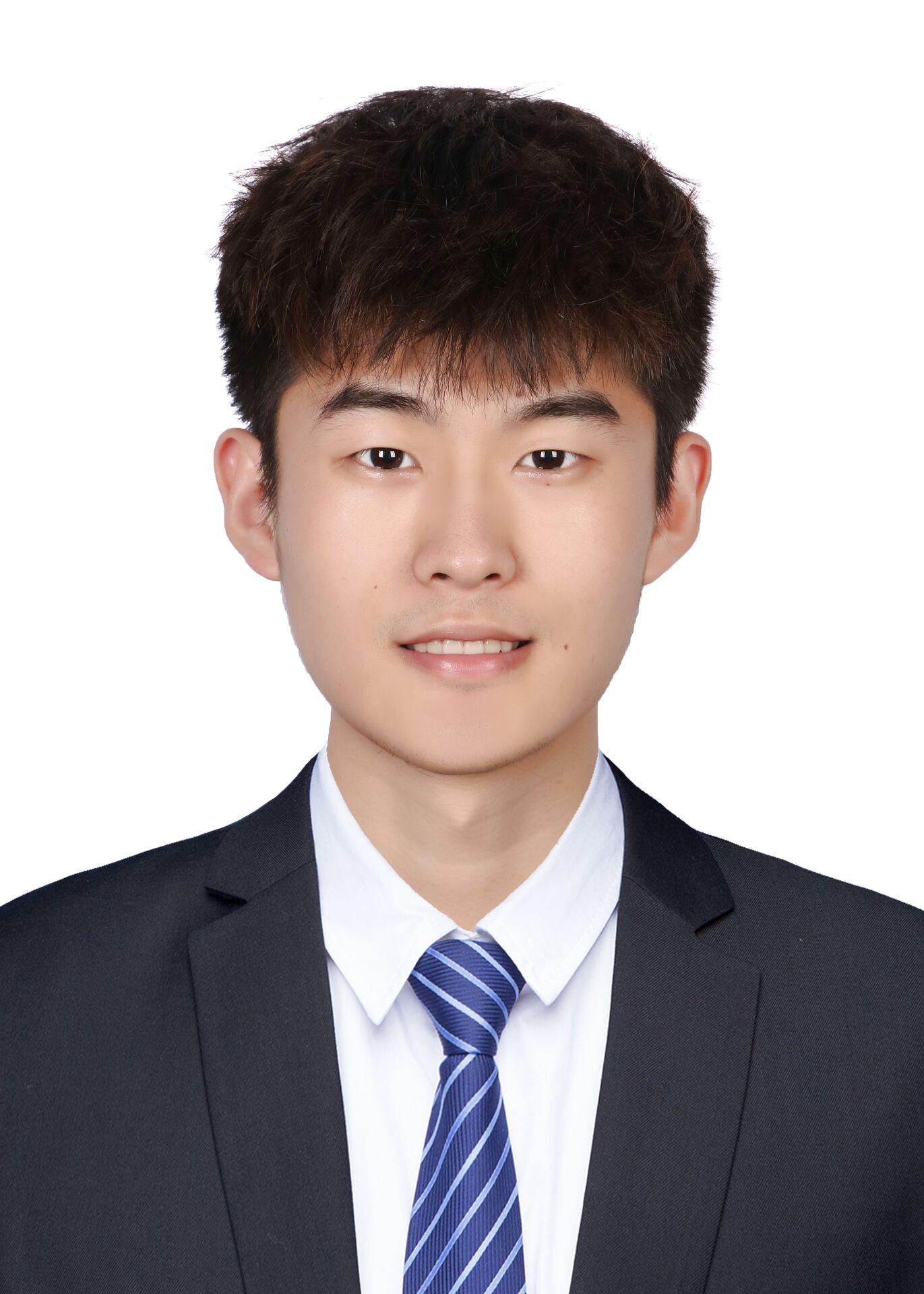}}]{Zhehan Li}
  received the B.E. degree from Xidian University, Xi'an, China, in 2022. He is currently a Ph.D. student at the School of Artificial Intelligence, Xidian University. His research interests include robotics and unmanned systems.
\end{IEEEbiography}

\begin{IEEEbiography}[{\includegraphics[width=1in,height=1.25in,clip,keepaspectratio]{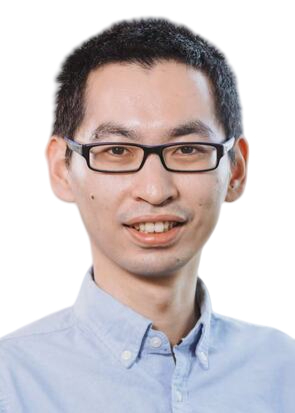}}]{Hangxin Liu}
    (Member, IEEE) received his Ph.D. degree in Computer Science from the University of California, Los Angeles (UCLA) in 2021. He is currently the leader of the robotics lab and a research scientist at State Key Laboratory of General Artificial Intelligence, Beijing Institute for General Artificial Intelligence (BIGAI). He received an M.S. degree in Mechanical Engineering from UCLA in 2018 and two B.S. degrees in Mechanical Engineering and Computer Science, both from Virginia Tech in 2016. His research interests focus on robot perception, learning, human-robot interaction, and cognitive robotics.
\end{IEEEbiography}

\begin{IEEEbiography}[{\includegraphics[width=1in,height=1.25in,clip,keepaspectratio]{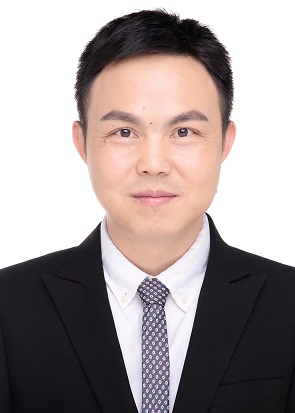}}]{Xinggang Wang}
  (Senior Member, IEEE) received the B.S. and Ph.D. degrees in Electronics and Information Engineering from Huazhong University of Science and Technology (HUST), Wuhan, China, in 2009 and 2014, respectively. He is currently a Professor at the School of Electronic Information and Communications, HUST. He serves as Co-Editor-in-Chief of Image and Vision Computing and area chair of CVPR and ICCV. His research interests include computer vision and deep learning.
\end{IEEEbiography}

\begin{IEEEbiography}[{\includegraphics[width=1in,height=1.25in,clip,keepaspectratio]{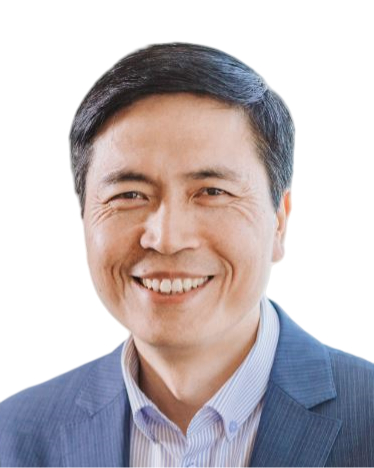}}]{Song-Chun Zhu}(Fellow, IEEE)
    received a Ph.D. degree from Harvard University in 1996, and is a Chair Professor jointly at Tsinghua University and Peking University, Dean of Institute for Artificial Intelligence at Peking University. He worked at Brown, Stanford, Ohio State, and UCLA before returning to China in 2020 to launch a non-profit organization---Beijing Institute for General Artificial Intelligence (BIGAI). He has published over 300 papers in computer vision, statistical modeling and learning, cognition, language, robotics, and AI. He received the Marr Prize in 2003, the Aggarwal prize from the Intl Association of Pattern Recognition in 2008, the Helmholtz Test-of-Time prize in 2013, twice Marr Prize honorary nominations in 1999 and 2007, the Sloan Fellowship, the US NSF Career Award, and the ONR Young Investigator Award in 2001. He served as General co-Chair for CVPR 2012 and CVPR 2019.
\end{IEEEbiography}




\end{document}